\documentclass[sn-mathphys,Numbered]{sn-jnl}

\usepackage{graphicx}%
\usepackage{multirow}%
\usepackage{amsmath,amssymb,amsfonts}%
\usepackage{amsthm}%
\usepackage{mathrsfs}%
\usepackage[title]{appendix}%
\usepackage{xcolor}%
\usepackage{textcomp}%
\usepackage{manyfoot}%
\usepackage{booktabs}%
\usepackage{algorithm}%
\usepackage{algpseudocode}%
\usepackage{listings}%
\usepackage{array}

\usepackage{algorithm}
\usepackage{algpseudocode}

\usepackage{verbatim}

\theoremstyle{thmstyleone}%
\newtheorem{theorem}{Theorem}
\newtheorem{proposition}[theorem]{Proposition}%

\theoremstyle{thmstyletwo}%
\newtheorem{example}{Example}%
\newtheorem{remark}{Remark}%

\theoremstyle{thmstylethree}%
\newtheorem{definition}{Definition}%

\begin{document}

\title[Towards Robust Ferrous Scrap Material Classification with Deep Learning and Conformal Prediction]{Towards Robust Ferrous Scrap Material Classification with Deep Learning and Conformal Prediction}


\author[1]{\fnm{Paulo Henrique} \sur{dos Santos}}

\author[2]{\fnm{Valéria de Carvalho} \sur{Santos}}

\author*[2]{\fnm{Eduardo José} \sur{da Silva Luz}}\email{eduluz@ufop.edu.br}


\affil*[1]{\orgdiv{Programa de Pós-Graduação em Instrumentação Controle e Automação de Processos de Mineração}, \orgname{Universidade Federal de Ouro Preto e Instituto Tecnológico Vale}}

\affil[2]{\orgdiv{Departamento de Computação}, \orgname{Universidade Federal de Ouro Preto}}



\abstract{In the steel production domain, recycling ferrous scrap is essential for environmental and economic sustainability, as it reduces both energy consumption and greenhouse gas emissions. However, the classification of scrap materials poses a significant challenge, requiring advancements in automation technology. Additionally, building trust among human operators is a major obstacle. Traditional approaches often fail to quantify uncertainty and lack clarity in model decision-making, which complicates acceptance. In this article, we describe how conformal prediction can be employed to quantify uncertainty and add robustness in scrap classification. We have adapted the Split Conformal Prediction technique to seamlessly integrate with state-of-the-art computer vision models, such as the Vision Transformer (ViT), Swin Transformer, and ResNet-50, while also incorporating Explainable Artificial Intelligence (XAI) methods. We evaluate the approach using a comprehensive dataset of 8147 images spanning nine ferrous scrap classes. The application of the Split Conformal Prediction method allowed for the quantification of each model's uncertainties, which enhanced the understanding of predictions and increased the reliability of the results. Specifically, the Swin Transformer model demonstrated more reliable outcomes than the others, as evidenced by its smaller average size of prediction sets and achieving an average classification accuracy exceeding 95\%. Furthermore, the Score-CAM method proved highly effective in clarifying visual features, significantly enhancing the explainability of the classification decisions.}

\maketitle

\section{Introduction}\label{sec1}

Recycling ferrous scrap is not only essential for steel production but also vital for environmental sustainability and economic efficiency. As the steel industry is a major contributor to global energy consumption and greenhouse gas emissions, accounting for approximately 7\% and 7-9\% respectively \cite{kim2022decarbonizing}, innovative recycling methods are paramount. Utilizing scrap in steel production promotes the principles of the circular economy, significantly reducing the environmental footprint of manufacturing \cite{companero2021circular}.

The primary technology for recycling scrap is the Electric Arc Furnace (EAF) method. This technique, which melts scrap along with other requisite materials, is notably energy-efficient as it primarily utilizes electrical energy converted into heat via an electric arc \cite{odenthal2018review}. However, the EAF process requires meticulous material selection and preparation due to the diverse quality and composition of scrap classes, which includes various metal alloys \cite{companero2021circular}.

Traditionally, scrap classification and separation have depended on manual processes, employing the subjective judgment of trained inspectors. This approach, while experienced, is fraught with potential errors and inefficiencies, particularly due to the visual similarities among different classes of scrap.

The limitations of traditional methods have spurred significant research into automated systems. Advances in computer vision and deep learning have recently made it possible to improve the accuracy and safety of these processes. For instance, early studies by Wieczorek et al.~\cite{wieczorek2008classification} and Baumert et al.~\cite{baumert2008automated} utilized image analysis techniques to categorize and assess steel scrap before it enters the EAF, showcasing the feasibility of these technologies.

Building on these foundations, more recent efforts have employed advanced machine learning models such as InceptionV3, ResNet50, Xception, and Faster-RCNN to enhance scrap classification's accuracy and efficiency. For example, Smirnov et al. \cite{smirnov2020machine, smirnov2021deep} and Qin et al. \cite{qin2021research} have applied deep learning classifiers to refine the categorization process, and Gao et al. \cite{gao2023rgb} have introduced an RGB-D-based system for 3D classification and thickness measurement of scrap materials, thereby improving precision.

Despite the technological advancements facilitated by the availability of public datasets, which have significantly enhanced the capabilities of both supervised and unsupervised learning methods, several challenges persist~\cite{schafer2023does}. These include the need for more comprehensive datasets that accurately reflect the diverse types of scrap materials encountered in industry and the adaptation of models to handle this variety effectively. However, one critical challenge is building operator trust in automated systems. Trust is essential for the adoption of new technologies, yet skepticism often exists among operators who are accustomed to traditional, manual processes. This skepticism is primarily due to concerns over the reliability and accuracy of automated methods in real-world, variable conditions. Addressing these concerns through transparent, understandable, and verifiable results is crucial to integrating advanced computational techniques into established industrial practices.

In this study, we introduce a statistical approach, as detailed in \cite{vovk1999machine, vovk2005algorithmic}, to calibrate the predictions of state-of-the-art deep learning architectures—specifically, the Vision Transformer (ViT) \cite{dosovitskiy2020image}, Swin Transformer \cite{DBLP:journals/corr/abs-2103-14030}, and the well-established ResNet-50 \cite{he2016deep}—for ferrous scrap classification. This approach is augmented with techniques from Explainable Artificial Intelligence (XAI). Our dual objectives are to enhance the accuracy and reliability of these models in industrial applications and to improve their interpretability.

Our research is driven by two principal questions: \textbf{How does integrating Conformal Prediction with advanced deep learning models improve the certainty and reliability of classification outcomes in industrial settings, specifically in ferrous scrap classification}? And, \textbf{How do various explainability methods clarify the decision-making processes of these models when classifying complex scrap material images, particularly those with atypical features}? To investigate these questions, we compiled a dataset of 8147 images across nine distinct classes of ferrous scrap. We evaluated three classifiers using multiple explainability methods, including Grad-CAM \cite{Selvaraju_2019}, Grad-CAM++ \cite{Chattopadhay_2018}, Score-CAM \cite{wang2020scorecam}, Eigen-CAM \cite{muhammad2020eigen}, and Deep Feature Factorization (DFF) \cite{collins2018deep}.

We adapted the Split Conformal Prediction method \cite{papadopoulos2002inductive, lei2018distribution} to assess model uncertainties and achieved classification accuracies of 95.00\% for ResNet-50, 95.15\% for ViT, and 95.51\% for Swin. This approach also yielded recall rates of 90.49\% for ResNet-50, and 95.12\% for both ViT and Swin, with Swin achieving the smallest prediction set size at 0.9878, indicating its superior reliability. Among the tested explainability methods, Score-CAM, when applied to the Swin model, proved most effective at illuminating key features through heat maps.

\section{Related Works}

This section provides a focused review of key studies concerning the automatic classification of ferrous and steel scrap, excluding research on other types of metal scrap. Historically, this field depended on manually crafted feature extraction methods. However, the landscape has shifted significantly due to recent advancements in deep learning techniques, which have markedly improved classification outcomes. Furthermore, the availability of open datasets has catalyzed further developments in the classification of steel and ferrous scrap, broadening the scope and applicability of these advanced methods.

In \cite{wieczorek2008classification}, the authors proposed a system for acquiring images of scraps loaded into an EAF. The images were captured when the electromagnet was carrying the scrap for deposition into the EAF. Computational vision methods were implemented to segment the scrap in the image. A BLOB algorithm was implemented to fragment and estimate the size of each object by identifying objects with similar intensities. Furthermore, parameters such as average and standard deviation were extracted from the scrap color values. This entire implementation of image processing was carried out to collect the mentioned data for building a database to develop a classifier model later. However, the sequence of this proposal with the classifier model was not found in the literature.

The authors in \cite{baumert2008automated} introduced a computational vision system with a laser scanner. This system is designed to accurately estimate the density of scrap materials, thereby enhancing the classification process. The scraps are loaded gradually into two baskets. A camera and laser scanner monitor each basket. As the basket is filled with scraps, the laser scanner obtains a superficial distribution of scraps and measures the height difference among the layers. This information is used to calculate the density. The image processing extracts characteristics from scraps, quantifying the granulometry and the distribution of the sizes of scraps. Then, this information is used in a classifier model of a probabilistic neural network that was trained to classify four classes. The results show an accuracy of 84,5\% for the inference of 200 images, 50 images for each class. However, neither the stratification of the dataset nor the training samples was presented.

Another proposal for ferrous scrap classification is presented in \cite{smirnov2020machine}. This work aimed to create a scrap classifier based on deep learning. First, a method was created for processing and preparing a dataset of scrap images in train carriages. The idea of the method was to cut only the scraps from the original photos and use them to train the model. Using a modest dataset divided into two, with approximately 120 images for each of nine classes and 200 images for each of four classes, the authors developed some classifiers modifying the InceptionV3, ResNet50, and Xception nets. Furthermore, they applied other machine learning methods, such as logistic regression, random forest, and gradient boosting. The best result was achieved by adapting InceptionV3 and ResNet50 for the four-class problem. For the nine-class problem, the modification of the Xception net had the best performance.

One year later, the same authors published a continuation of their work in \cite{smirnov2021deep}. In this study, they used a larger dataset, about 1080 images for each of the nine classes and 1200 images for each of the four classes. It used the pre-trained DenseNet201, InceptionResNetV2, ResNet152V2, and NASNetLarge networks. The modified network based on NASNetLarge showed better performance in classification for the nine-class case. In the four-class case, the networks based on ResNet152V2 and DesenNet201 presented better performance. The authors also proposed an isolated model for the binary classification of each class. In this manner, the result of each model determines whether the input image belongs to that respective class. The results showed that the binary models were not better than the multi-class classifiers for both four- and nine-class problems.

In \cite{qin2021research}, the use of a Faster-RCNN network for the classification and detection of ferrous scrap was proposed. This object detection model identifies the scrap position and class in the image. The dataset used had five classes, with 80 images for each class. The model reached 87,6\% of accuracy.

In \cite{armellini2022q}, the authors described the Q-SYM2, a scrap management system projected for optimizing the flow from the beginning of the chain to its use in EAF. The system employs computational vision tools to classify scrap, estimate its actual weight, and identify discrepancies in the theoretical weight. A crucial stage of Q-SYM2 is automatically classifying scraps and detecting dangerous materials. The neural network model used presented 95\% confidence in classification. According to the authors, this allows the complete elimination of manual classification in the scrap receiving process. However, the paper must provide details about the deep learning model or dataset. Furthermore, the 95\% mentioned represents the confidence of the model outputs, not the accuracy. The authors affirm that training of the model was still in progress for the accuracy improvement, which was not presented.

In contrast to other proposals that use 2-D images as input, the authors in \cite{gao2023rgb} proposed the use of a 3-D-based system, which provides the estimation of not only classes but also the thickness of scraps. For this purpose, they created a public dataset with RGB-D images of 29 distinct categories, summing up 3440 labeled images and 6081 samples. Using an approach based on a point cloud, the trained model is capable of calculating and estimating the distribution of scrap thickness, using this information to identify the class and localization in the corresponding image, working as an object detector. The authors compared their model with the Faster-RCNN \cite{qin2021research}, YOLOX \cite{ge2021yolox} and DETR \cite{carion2020end}, revealing that the model presented an average precision of 80,1\%, as Faster-RCNN has reached 34,21\%, YOLOX 54,55\% and DETR 28,21\%.

Another study that provides an open and public dataset is presented in \cite{schafer2023does}. The researchers collected videos from the piles of scraps in a scrap yard, which were processed to convert the videos into images. All images were processed and labeled, and the background was separated. Furthermore, the images were cropped to obtain better performance, resulting in a dataset with RGB images of 256 x 256 pixels. The dataset consists of 6020 raw images and 105523 processed and cropped images for training. For the tests, there were 176 raw images and 8131 processed and cropped images. All these images are distributed in nine classes. To validate the dataset, the authors trained the PreActResNet18 \cite{he2016identity} and ResNet50 \cite{he2016deep} networks with both raw and processed images. The PreActResNet18 network obtained the best result, reaching 68,09\% of accuracy in the training with the cropped images and 43,18\% with the raw images.

This review highlights that, although numerous studies on ferrous and steel scrap classification have yielded promising results, they frequently lack in areas such as explainability and the quantification of uncertainty or non-conformity in their findings. Typically, these studies prioritize achieving high accuracy while neglecting to provide transparency regarding the crucial features that drive class distinctions. However, explainability and uncertainty quantification are essential for the practical implementation of these technologies in industrial settings, as they boost user confidence and facilitate broader acceptance. Therefore, this work aims not only to enhance classification accuracy but also to significantly improve the methods' explainability and their ability to quantify uncertainties.

\section{Materials and methods}\label{sec2}

This section outlines the dataset composition, scrap classification approach, deep learning architectures, conformal prediction methodologies, and explainability techniques used in our study. The investigation seeks to answer two research questions posed in this study.


\subsection{The data}

\begin{figure}[h]%
\centering
\includegraphics[width=0.3\textwidth]{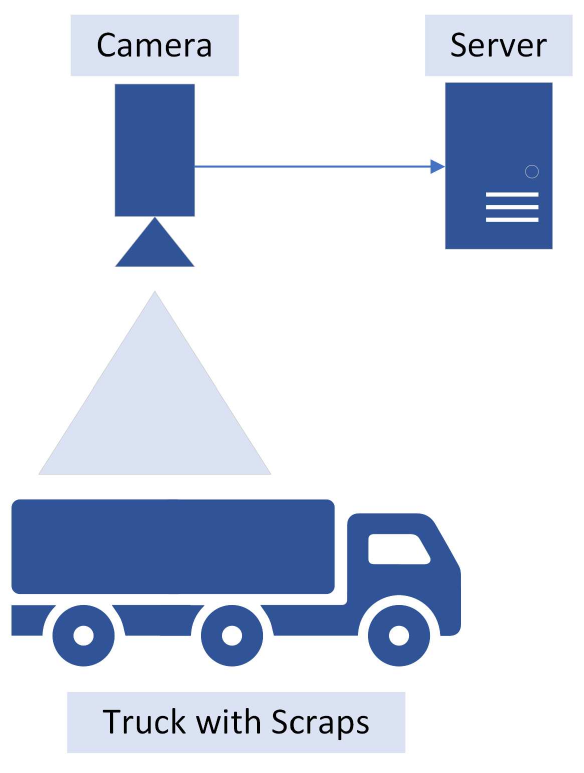}
\caption{Scheme for automatic image capture with a fixed camera.}\label{fig:camera}
\end{figure}

The dataset comprises images captured by a stationary camera with a resolution of 1456 x 1088 pixels. This camera was positioned to photograph the truck beds upon their arrival at the steel manufacturing facility, as illustrated in Figure \ref{fig:camera}. The camera captured images over a duration of 115 days, spanning from June to September 2023. The dataset underwent a detailed curation process by domain experts, resulting in a total of 8147 valid samples being compiled (See Table \ref{tab:dataset}).

\begin{figure}[h]
  \centering
  \begin{minipage}{0.3\textwidth}
    \centering
    Steel Sheets\\
    \includegraphics[width=\linewidth]{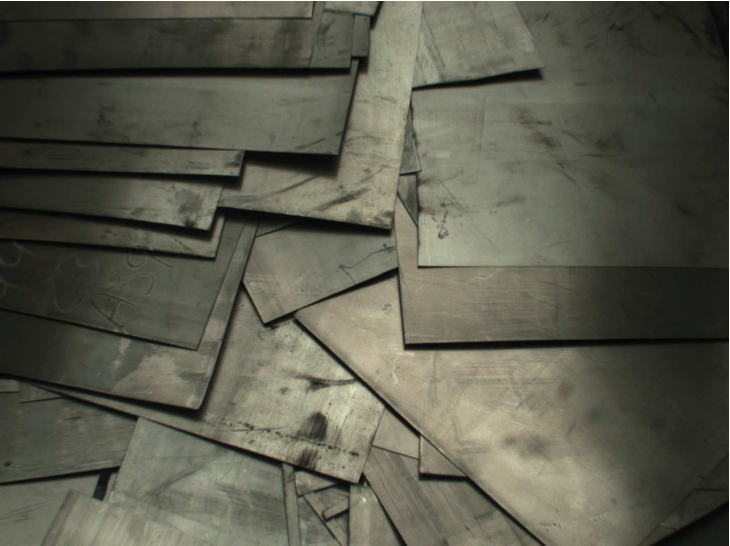}
  \end{minipage}
  \begin{minipage}{0.3\textwidth}
    \centering
    Stamping Scrap\\
    \includegraphics[width=\linewidth]{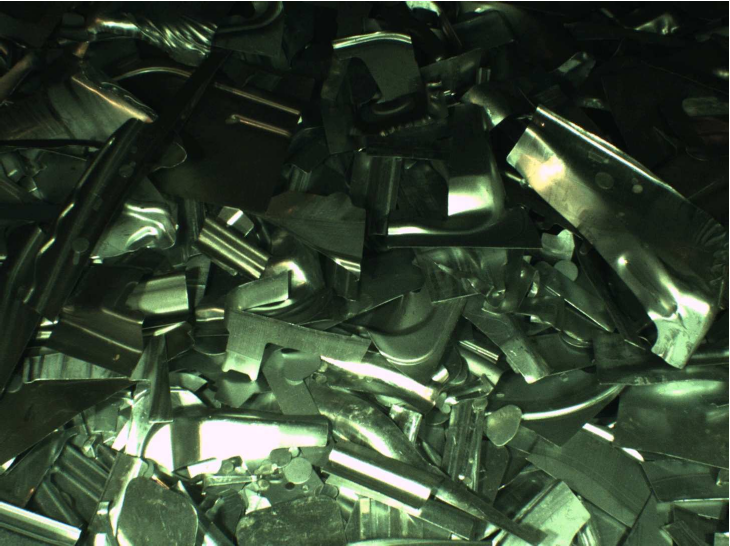}
  \end{minipage}
  \begin{minipage}{0.3\textwidth}
    \centering
    Swarf Scrap\\
    \includegraphics[width=\linewidth]{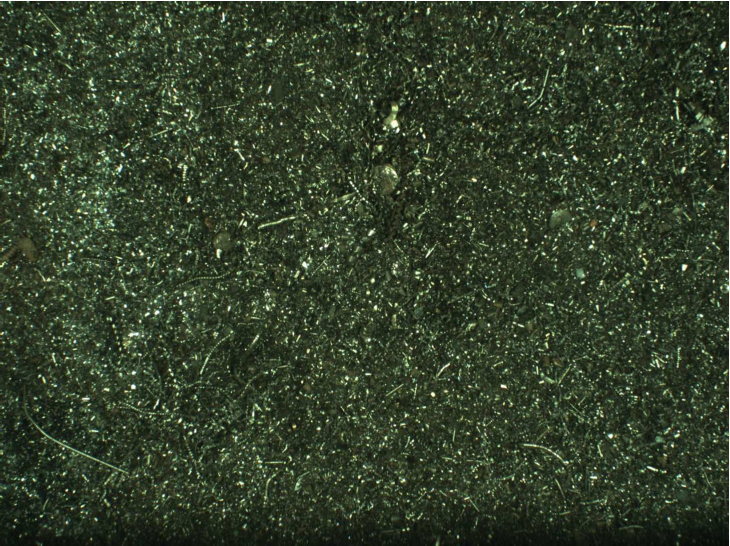}
  \end{minipage}
  
  
  \begin{minipage}{0.3\textwidth}
    \centering
    High-Quality Oxyfuel Cutting\\
    \includegraphics[width=\linewidth]{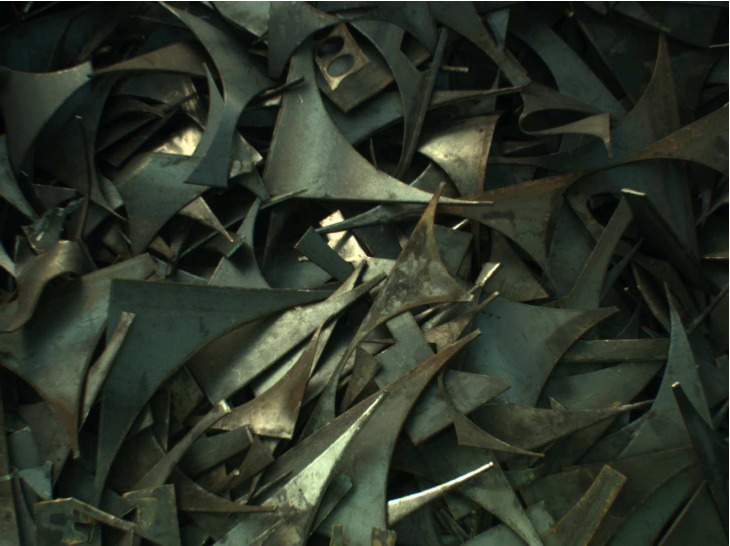}
  \end{minipage}
  \begin{minipage}{0.3\textwidth}
    \centering
     Low-Quality Oxyfuel Cutting\\
    \includegraphics[width=\linewidth]{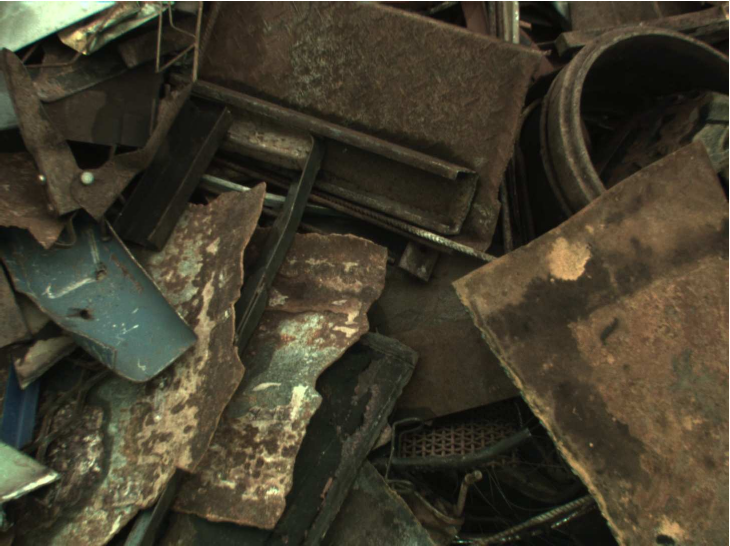}
  \end{minipage}
  \begin{minipage}{0.3\textwidth}
    \centering
    Shredder\\ \bigskip
    \includegraphics[width=\linewidth]{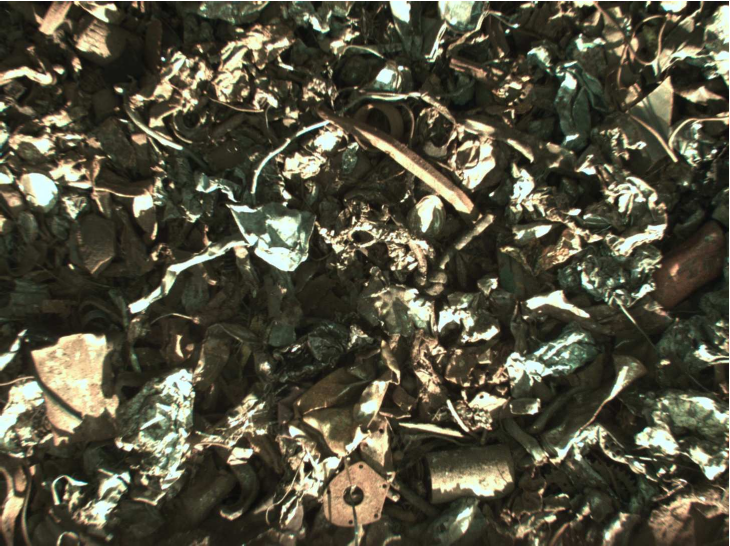}
  \end{minipage}
  
  
  \begin{minipage}{0.3\textwidth}
    \centering
    Sheared Scrap\\
    \includegraphics[width=\linewidth]{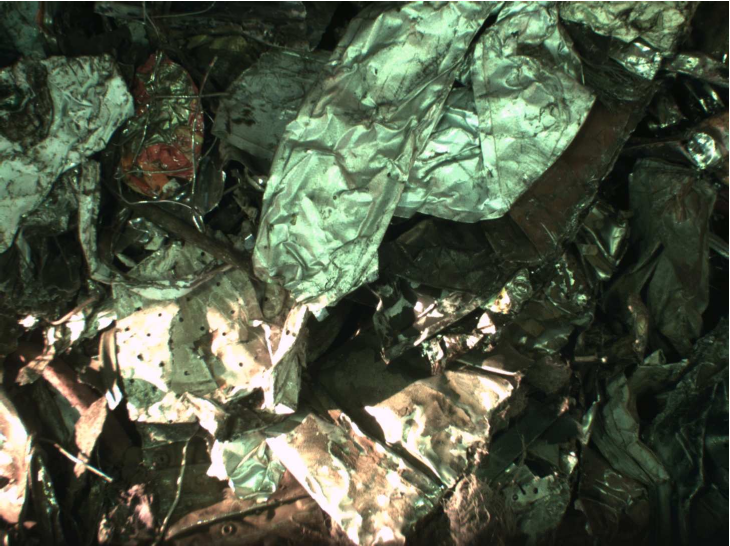}
  \end{minipage}
  \begin{minipage}{0.3\textwidth}
    \centering
    High-Quality Packages\\
    \includegraphics[width=\linewidth]{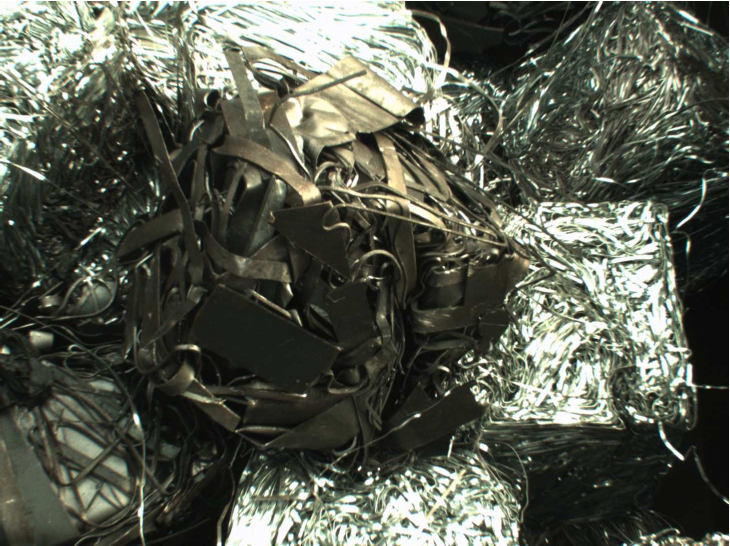}
  \end{minipage}
  \begin{minipage}{0.3\textwidth}
    \centering
    Low-Quality Packages\\
    \includegraphics[width=\linewidth]{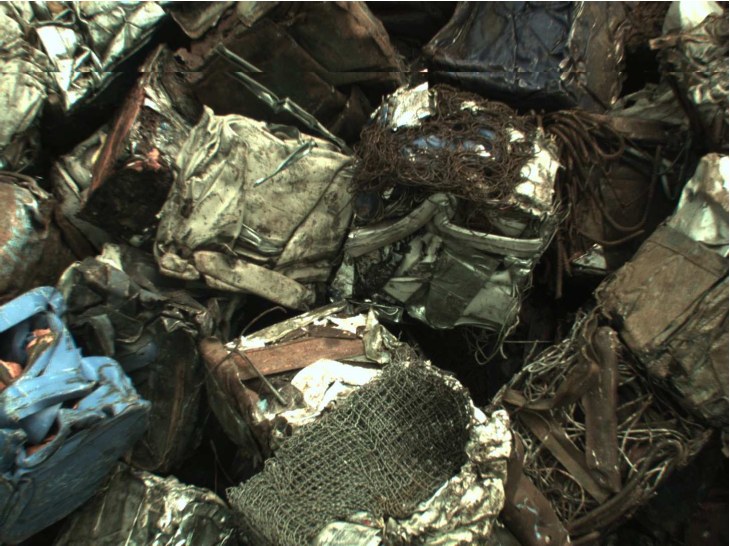}
  \end{minipage}
  
  \caption{Image example of each class.}\label{fig:classes}
\end{figure}

In this research, nine distinct classes of scrap metal are examined, ranging from high-quality, which are largely free of contaminants or contain minimal amounts of non-metal materials like ink, plastic, and rubber, to those considered low-quality due to the presence of such contaminants. The following section provides a detailed description of each class, highlighting the predominant characteristics observed in the dataset for each category. Figure \ref{fig:classes} presents a representative example from each class, while Table \ref{tab:dataset} outlines the distribution of samples across the various classes.

\begin{enumerate}[1.]
\item Steel Sheets: This category encompasses remnants from steel sheets tailored for manufacturing a variety of products, predominantly within the automotive sector \cite{matlock2012recent}. Such scrap is considered to be high-quality, free from contaminants, and exhibits a metallic gray hue.

\item Stamping Scrap: Derived from stamping or pressing processes, this scrap is shaped, cut, or molded using stamping machines \cite{karbasian2010review}. It is recognized for its high quality, absence of contaminants, compact size, and metallic gray appearance.

\item Swarf Scrap: Generated during machining operations, swarf scrap consists of fine metal fragments, chips, shavings, or filings \cite{kopac2009concepts}. This lower-quality scrap is characterized by small, uniform pieces and a darker gray coloration.

\item High-Quality Oxyfuel Cutting Scrap: This high-quality, contaminant-free scrap is produced by the oxyfuel cutting process \cite{kulkarni2008micro}. Sourced directly from industrial outputs, it features uniform sizes and primarily comprises meticulously cut small metal pieces.

\item Low-Quality Oxyfuel Cutting Scrap: Originating from larger iron scrap pieces replete with contaminants, this class is formed by applying the oxyfuel cutting technique to materials gathered from scrapyards or natural settings. It displays variable sizes and may include irregular metal objects such as car wheels, bars, and cylinders, among others, marked by the presence of contaminants and size diversity.

\item Shredder Scrap: Produced through shredding operations, where machines employ rotating hammers to fragment scrap into finer pieces \cite{russo2004mechanical}. This process segregates metal from non-metal component post-shredding, thus, this class predominantly consists of small metal fragments with ink as the main contaminant.

\item Sheared Scrap: Closely mirroring the Low-Quality Oxyfuel Cutting Scrap, this category uses the same raw materials but employs shear machines for the cutting process \cite{karagulle2019multi}. It contains contaminants, exhibits significant size variation, and the materials often appear crumpled.

\item High-Quality Packages: Comprising materials akin to Stamping Scrap or High-Quality Oxyfuel Cutting Scrap, this class is processed and configured into cube-shaped packages or bundles.

\item Low-Quality Packages: Constituted of materials comparable to Sheared Scrap but processed into cube-shaped packages or bundles, representing the lower-quality spectrum.

\end{enumerate}

\begin{table}[h]
  \centering  
  \caption{Dataset classes distribution}\label{tab:dataset}
  \begin{tabular}{@{}ll@{}}
    \toprule
    Class & Samples \\
    \midrule
    Steel Sheets & 296 \\
    Stamping Scrap & 1490 \\
    Swarf Scrap & 1141 \\
    High-Quality Oxyfuel Cutting Scrap & 1170 \\
    Low-Quality Oxyfuel Cutting Scrap & 865\\
    Shredder & 372 \\
    Sheared Scrap & 1180 \\
    High-Quality Packages & 786 \\
    Low-Quality Packages & 847 \\
    \midrule
    \textbf{TOTAL} & 8147 \\
    \bottomrule
  \end{tabular}
\end{table}

\subsection{Deep Learning Models}

Aiming automatic steel scrap classification, this study evaluates three distinct deep learning architectures: ResNet-50 \cite{he2016deep}, the base version of the Vision Transformer (ViT) \cite{dosovitskiy2020image}, and the tiny version of the Swin Transformer \cite{DBLP:journals/corr/abs-2103-14030}. Employing a transfer learning approach, we leveraged pre-existing knowledge from datasets to speed the learning process for our specific dataset \cite{pan2009survey}, and enhance model training efficiency. Specifically, the ViT model underwent pre-training on the ImageNet-21k dataset \cite{deng2009imagenet}, while ResNet-50 and Swin Transformer were initialized with weights from training on ImageNet-1k \cite{deng2009imagenet}, all adapted to a resolution of 224 x 224 pixels. Subsequently, each model was fine-tuned using images from our dataset, as detailed in Table \ref{tab:dataset}, with necessary adjustments to resize each image to 224 × 224 pixels for model input compatibility. 

The dataset was divided into three distinct segments: 75\% allocated for training, 15\% designated for validation, and the remaining 10\% reserved for testing. The training and validation partitions are used throughout the model training phase to adjust and evaluate the model's performance iteratively. Subsequently, the test partition is employed exclusively post-training to assess the model's efficacy on previously unseen data, thereby ensuring an unbiased evaluation of its generalization capabilities.

To ensure the reliability and statistical significance of our results, each model underwent training ten times each with a different random seed, with the dataset being shuffled before each iteration while maintaining the established partition percentages. The training hyperparameters, including the number of epochs, steps per epoch, and learning rate, among others, were consistently held constant across all models. This methodological rigor was applied to guarantee the comparability and reproducibility of the findings, facilitating an accurate assessment of each model's performance.

\subsection{Conformal Prediction Approach}\label{sec_methods_conformal}

Following the training phase, each model is subjected to the Split Conformal Prediction process \cite{papadopoulos2002inductive, lei2018distribution}. This necessitates dividing the original test dataset into a new test set and a calibration set. The calibration set is instrumental in identifying a threshold to ensure that the classification model achieves a predetermined coverage rate of 95\%, as formalized in Equation \ref{eq:coverage}:
\begin{equation}\label{eq:coverage}
\textit{coverage} = 1 - \alpha,
\end{equation}
where $\alpha \in (0,1)$ indicates the miscoverage rate. Setting $\alpha$ to 0.05 targets the desired 95\% coverage. This methodical partitioning of the dataset and the calibration of coverage levels are critical for evaluating the model's generalization capabilities with a defined accuracy level.

Determining the calibration threshold involves running inferences for each sample in the calibration set and computing the non-conformity score $s_i$, as shown in:
\begin{equation}\label{eq:si}
s_i = 1 - \textit{predicted true class probability},
\end{equation}
which leverages the predicted probability of the actual class labeled in the sample. This yields an array of $s_i$ scores for the calibration samples.

The quantile level $qlevel$ is calculated using:
\begin{equation}\label{eq:qlevel}
qlevel = 0.95(n+1)/n,
\end{equation}
where $n$ is the number of calibration samples and $(n+1)/n$ applies a finite sample correction. The calibration threshold is identified by locating the 95th percentile of the $s_i$ scores, indicating that 95\% of the $s_i$ values are equal to or lower than this threshold. This calibration procedure, detailed in Algorithm \ref{algorithm:calibration}, provides a structured method for establishing the calibration threshold according to the specified coverage level, ensuring a more objective assessment of model performance.

\begin{algorithm}
\caption{Calibration for Split Conformal Prediction}\label{algorithm:calibration}
\begin{algorithmic}[1]

\State \textbf{Input}: Calibration dataset $D_{\text{calib}}$, Model $M$, Miscoverage level $\alpha$
\State \textbf{Output}: Threshold $\tau$

\Procedure{Calibration}{$D_{\text{calib}}, M, \alpha$}
    \State Initialize an empty list $S$
    \For{each sample $x_i$ in $D_{\text{calib}}$}
        \State Calculate $s_i = 1 - \textit{predicted true class probability for } x_i \textit{ by } M$
        \State Add $s_i$ to list $S$
    \EndFor
    \State Sort list $S$ in ascending order
    \State Calculate $qlevel = (1 - \alpha) \cdot (\lvert D_{\text{calib}} \rvert + 1) / \lvert D_{\text{calib}} \rvert$
    \State Set threshold index $idx = \lceil qlevel \cdot \lvert S \rvert \rceil$
    \State Set threshold $\tau = S[idx]$
    \State \textbf{return} $\tau$
\EndProcedure

\end{algorithmic}
\end{algorithm}

Therefore, rather than determining the predicted class solely on the basis of the maximum probability output by the model, it is feasible to compute the non-conformity score $s_i$ for each output probability and assess whether this score exceeds the established threshold. If so, the corresponding class is included in the prediction set. Consequently, for each inference, the conformal prediction yields a set containing one or more classes that have a 95\% likelihood of representing the true outcome, as calibrated against the dataset. This approach allows for a more probabilistically grounded classification. The process of performing inferences on the new test partition using this method is detailed in Algorithm \ref{algorithm:inference}.
 
\begin{algorithm}
\caption{Inference for Split Conformal Prediction}\label{algorithm:inference}
\begin{algorithmic}[1]

\State \textbf{Input}: New test dataset $D_{\text{new\_test}}$, Model $M$, Threshold $\tau$
\State \textbf{Output}: Predictions for $D_{\text{new\_test}}$

\Procedure{Inference}{$D_{\text{new\_test}}, M, \tau$}
    \State Initialize an empty list $P$ to store predictions for each sample
    \For{each sample $x_j$ in $D_{\text{new\_test}}$}
        \State Initialize an empty set $S_{x_j}$ to store predictions for sample $x_j$
        \For{each possible class $c_k$}
            \State Calculate $s_{j,k} = 1 - \textit{probability of } c_k \textit{ for } x_j \textit{ by } M$
            \If{$s_{j,k} > \tau$}
                \State Add $c_k$ to the prediction set $S_{x_j}$ for sample $x_j$
            \EndIf
        \EndFor
        \State Add $S_{x_j}$ to the list $P$
    \EndFor
    \State \textbf{return} $P$
\EndProcedure

\end{algorithmic}
\end{algorithm}

\subsection{Explainability}

In this study, we employ various explainability techniques to elucidate the decision-making mechanisms of Deep Learning models tasked with classifying steel scrap. The aim is to visually discern the focus areas of the model when making classification decisions. Techniques such as Grad-CAM \cite{Selvaraju_2019}, Grad-CAM++ \cite{Chattopadhay_2018}, Eigen-CAM \cite{muhammad2020eigen}, Score-CAM \cite{wang2020scorecam}, and Deep Feature Factorization (DFF) \cite{collins2018deep} are applied.

The selection of a target layer within the Deep Learning model for generating heat maps, which indicate the regions most critical to the model’s classification, involves empirical testing to identify the most informative layer. For the ResNet-50 model, the final encoder layer is chosen for the application of Grad-CAM, Grad-CAM++, Eigen-CAM, Score-CAM, and DFF methods. For the Vision Transformer (ViT) model, the penultimate encoder layer output serves as the target layer for Grad-CAM, Grad-CAM++, Eigen-CAM, and Score-CAM, while the layernorm layer is selected for DFF application with ViT. For the Swin Transformer model, the layernorm layer is consistently selected across all the abovementioned explainability techniques.

Deep Feature Factorization (DFF) is an innovative method that segments an image into distinct sections and determines the class of each segment through model inference. The decision on how many segments to divide the image into is made empirically, opting to split the image into five parts for this analysis. This segmentation facilitates a comprehensive examination of how each part of the image contributes to the model's classification decision, offering insights into the model's interpretative process.

\section{Results}\label{sec3}

This section presents the experimental outcomes that are structured to address the research questions posited in this study. The experiments were developed with Pytorch library \cite{paszke2019pytorch} and \textit{The PyTorch library for CAM methods} \cite{jacobgilpytorchcam}. For training, an AWS Sage Maker instance \textit{ml.g4dn.2xlarge} (8 vCPU + 32GiB + 1 GPU) \cite{mishra2019machine} was used.

\subsection{Experiment \#1: Scrap Classification with Deep Learning Models}

Before addressing the research questions presented in this study, it is essential to evaluate state-of-the-art computer vision architectures to establish their performance benchmarks. This initial examination leads to the selection of a single model from each architecture for a more detailed and focused analysis.

To ensure robustness and reliability of the results, each model underwent training ten times with the dataset being shuffled before each iteration, while maintaining control over the random seed. The performance metrics for all three models, as detailed in Tables \ref{tab:resnet50_training}, \ref{tab:vit_training}, and \ref{tab:swin_training}, reveal an average test accuracy exceeding 95\%. Notably, the Swin Transformer model not only trained more rapidly but also achieved the highest test accuracy coupled with the smallest standard deviation among the models evaluated.

\begin{table}[h]
   \centering 
  \caption{Performance training metrics for Resnet-50}\label{tab:resnet50_training}
  \begin{tabular}{@{}>{\centering\arraybackslash}p{4.5cm} >{\centering\arraybackslash}p{2.5cm} >{\centering\arraybackslash}p{2.5cm} >{\centering\arraybackslash}p{2.5cm}@{}}
    \toprule
    \textbf{Training} & \textbf{Training Time(s)} & \textbf{Validation Accuracy} & \textbf{Test Accuracy} \\
    \midrule
    1 & 3992.00 & 0.9484 & 0.9485 \\
    2 & 3989.42 & 0.9566 & 0.9387 \\
    3 & 4035.47 & 0.9476 & 0.9534 \\
    4 & 3989.05 & 0.9509 & 0.9313 \\
    5 & 4039.82 & 0.9542 & 0.9571 \\
    6 & 3996.68 & 0.9599 & 0.9571 \\
    7 & 4074.22 & 0.9566 & 0.9436 \\
    8 & 4024.51 & 0.9444 & 0.9620 \\
    9 & 4051.66 & 0.9493 & 0.9521 \\
    10 & 4019.01 & 0.9483 & 0.9562 \\
    \midrule
    \textbf{Average} & 4021.18 & 0.9516 & 0.9500 \\
    \textbf{Standard deviation} & 27.93 & 0.0047 & 0.0091 \\
    \bottomrule
  \end{tabular}
\end{table}

\begin{table}[h]
  \centering 
  \caption{Performance training metrics for ViT}\label{tab:vit_training}
  \begin{tabular}{@{}>{\centering\arraybackslash}p{4.5cm} >{\centering\arraybackslash}p{2.5cm} >{\centering\arraybackslash}p{2.5cm} >{\centering\arraybackslash}p{2.5cm}@{}}
    \toprule
    \textbf{Training} & \textbf{Training Time(s)} & \textbf{Validation Accuracy} & \textbf{Test Accuracy} \\
    \midrule
    1 & 4349.88 & 0.9425 & 0.9526 \\
    2 & 4441.62 & 0.9550 & 0.9583 \\
    3 & 4263.74 & 0.9558 & 0.9497 \\
    4 & 4256.82 & 0.9542 & 0.9497 \\
    5 & 4247.57 & 0.9493 & 0.9448 \\
    6 & 4230.29 & 0.9517 & 0.9595 \\
    7 & 4317.89 & 0.9550 & 0.9460 \\
    8 & 4295.72 & 0.9517 & 0.9448 \\
    9 & 4305.14 & 0.9534 & 0.9620 \\
    10 & 4261.68 & 0.9558 & 0.9472 \\
    \midrule
    \textbf{Average} & 4297.04 & 0.9524 & 0.9515 \\
    \textbf{Standard deviation} & 59.23 & 0.0039 & 0.0060 \\
    \bottomrule
  \end{tabular}
\end{table}

\begin{table}[h]
  \centering 
  \caption{Performance training metrics for Swin}\label{tab:swin_training}
  \begin{tabular}{@{}>{\centering\arraybackslash}p{4.5cm} >{\centering\arraybackslash}p{2.5cm} >{\centering\arraybackslash}p{2.5cm} >{\centering\arraybackslash}p{2.5cm}@{}}
    \toprule
    \textbf{Training} & \textbf{Training Time(s)} & \textbf{Validation Accuracy} & \textbf{Test Accuracy} \\
    \midrule
    1 & 3131.36 & 0.9689 & 0.9485 \\
    2 & 3202.46 & 0.9664 & 0.9632 \\
    3 & 3224.14 & 0.9607 & 0.9521 \\
    4 & 3124.94 & 0.9591 & 0.9571 \\
    5 & 3236.11 & 0.9648 & 0.9534 \\
    6 & 3123.35 & 0.9566 & 0.9558 \\
    7 & 3166.49 & 0.9689 & 0.9546 \\
    8 & 3140.04 & 0.9624 & 0.9632 \\
    9 & 3156.00 & 0.9615 & 0.9546 \\
    10 & 3214.83 & 0.9639 & 0.9489 \\
    \midrule
    \textbf{Average} & 3171.97 & 0.9633 & 0.9551 \\
    \textbf{Standard deviation} & 41.39 & 0.0039 & 0.0048 \\
    \bottomrule
  \end{tabular}
\end{table}

\begin{figure}[h]
  \centering
  \begin{minipage}{0.4\textwidth}
    \centering
    \includegraphics[width=\linewidth]{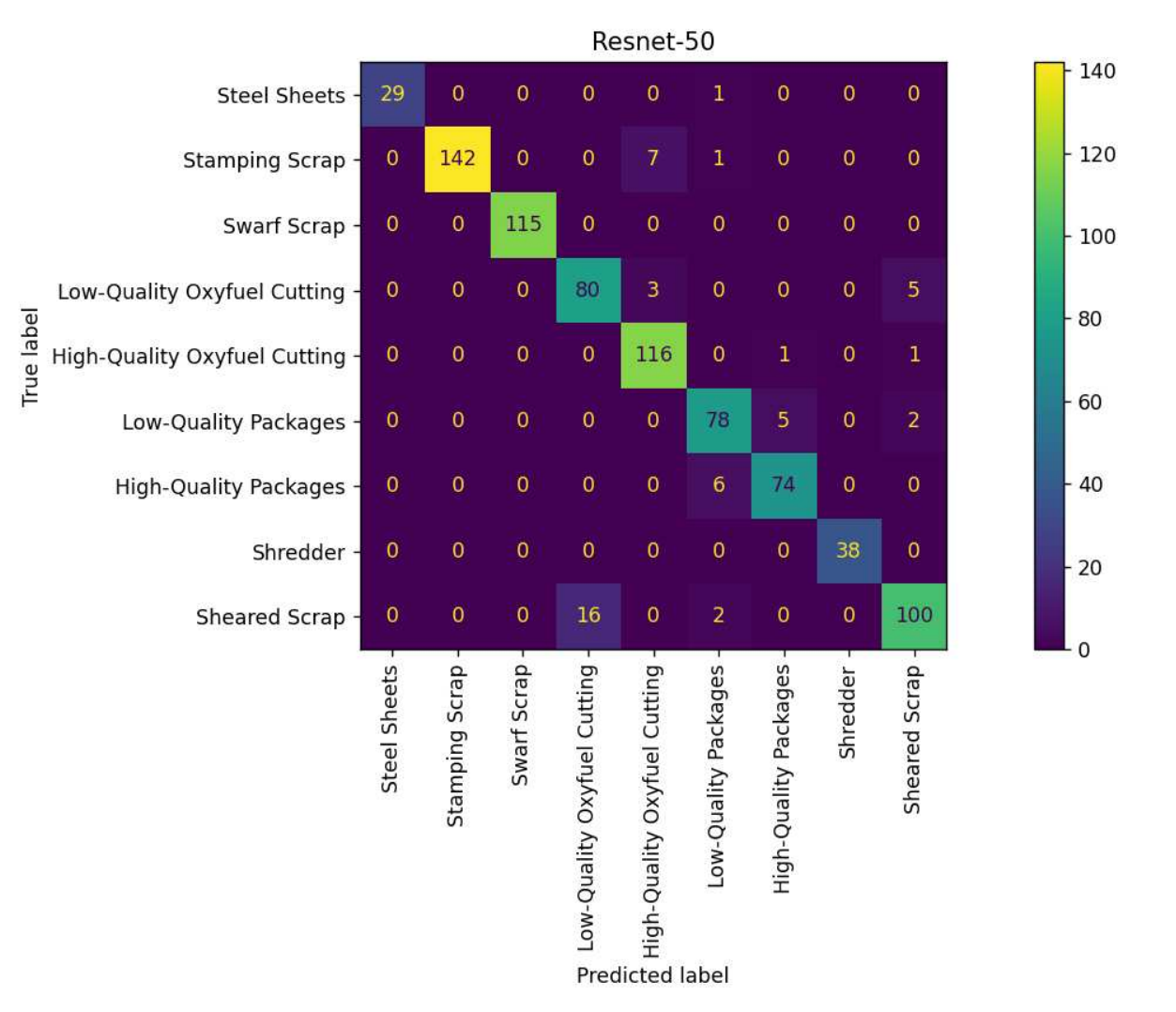}
  \end{minipage}
  \begin{minipage}{0.4\textwidth}
    \centering
    \includegraphics[width=\linewidth]{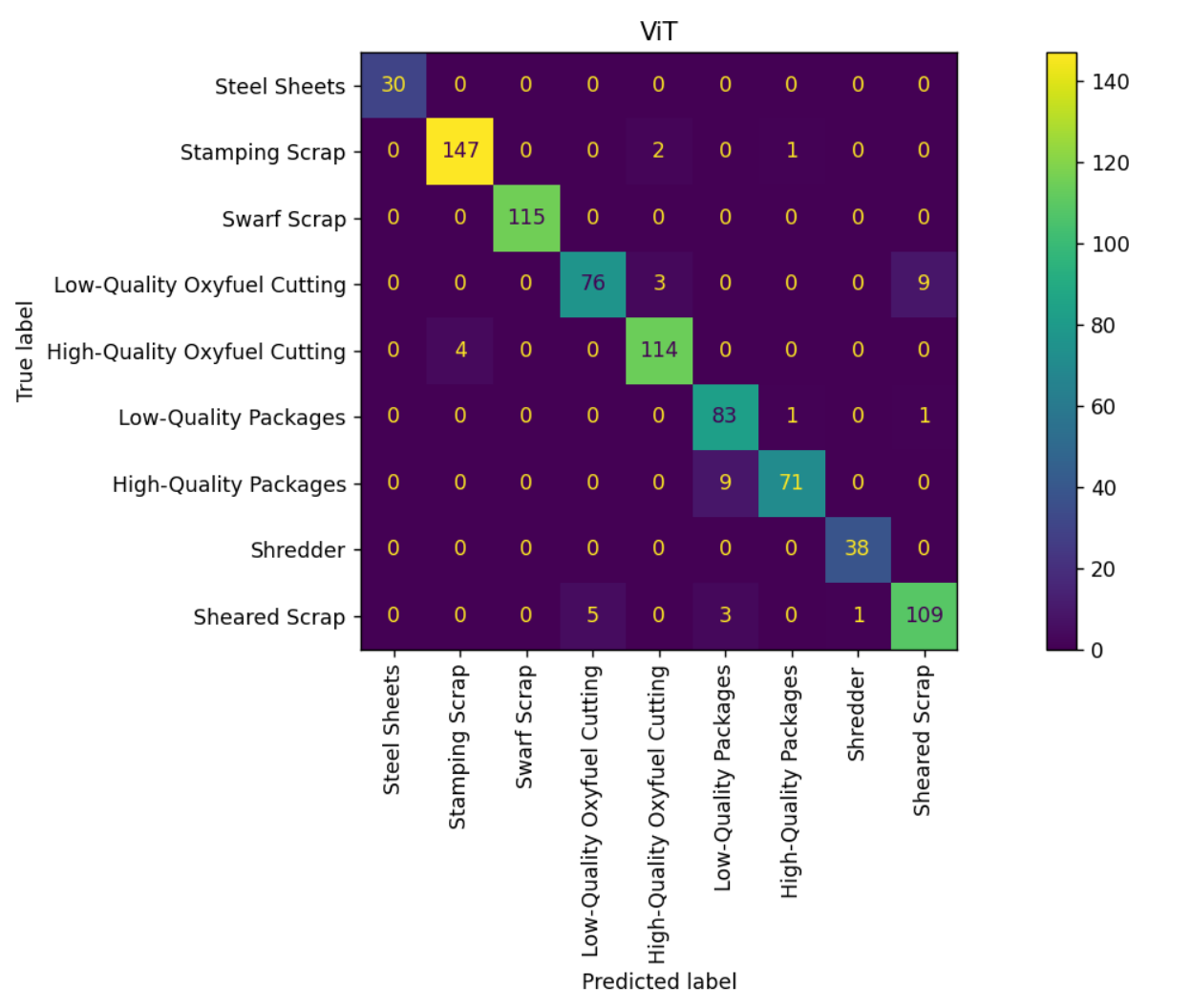}
  \end{minipage}
  \begin{minipage}{0.4\textwidth}
    \centering
    \includegraphics[width=\linewidth]{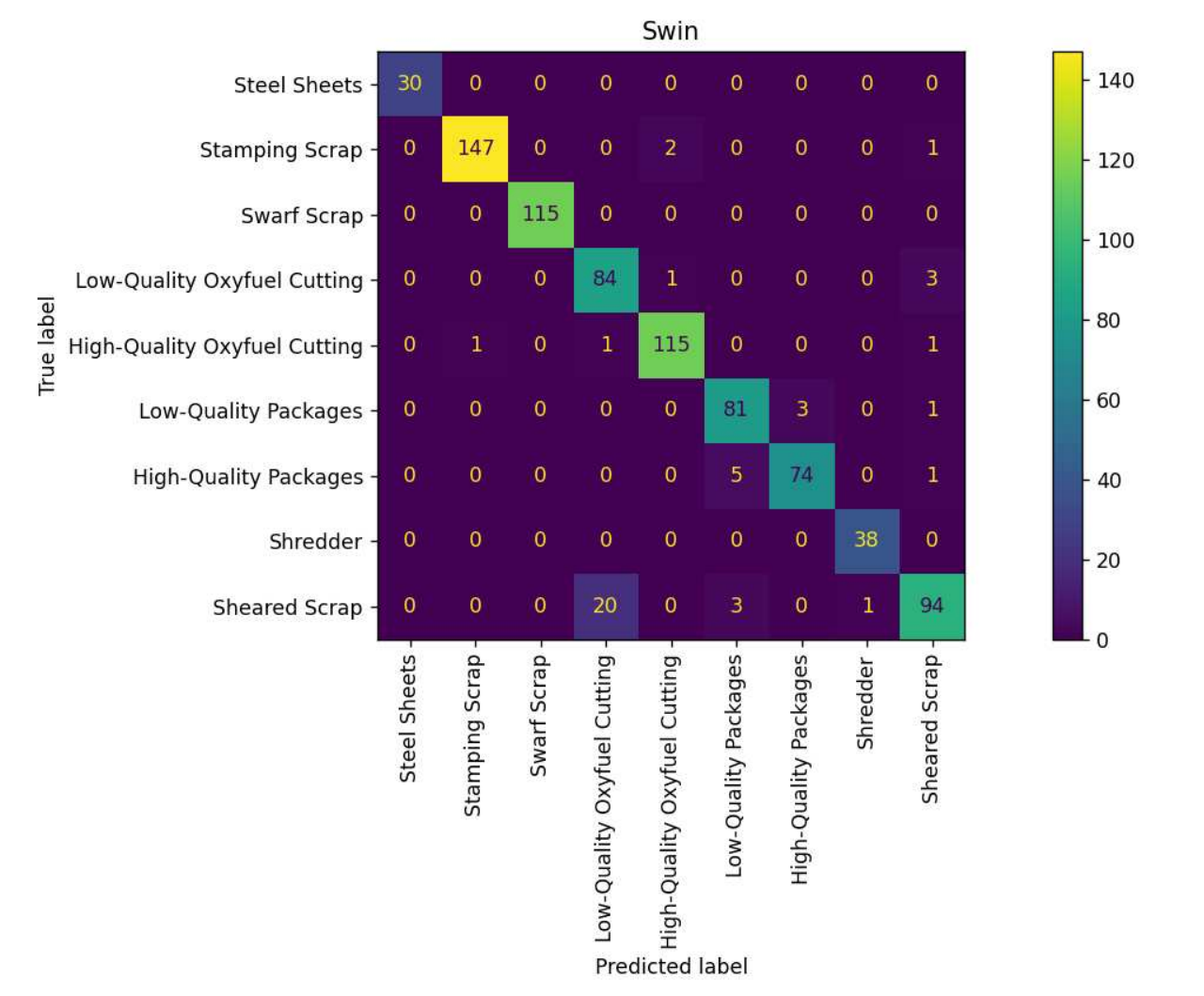}
  \end{minipage}
  
  \caption{Confusion Matrix of Resnet-50, ViT and Swin.}\label{fig:confusion_matrix}
\end{figure}

During each training cycle, attention was paid to the generation of mini-batches, ensuring uniformity in model training procedures. For subsequent experiments involving conformal prediction and explainability analyses, a singular version of each model—specifically, the initial model from each training set—was selected for these analyses. This strategy was employed to provide a stable and consistent framework for a detailed examination of these essential dimensions of model performance. The test dataset, which encompassed 822 images, served as the basis for assessing the effectiveness of the models. Figure \ref{fig:confusion_matrix} displays the Confusion Matrix for these selected versions of each model.

Table \ref{tab:recall} facilitates a comparison of recall for each class across different models. This table, along with the confusion matrices, illustrates that the Swin model exhibits superior recall across all classes except for Sheared Scrap and Low-Quality Packages. It further reveals that while all three models demonstrate competitive performance, the Vision Transformer (ViT) and Swin models stand out as the most effective. Moreover, the average accuracy across different versions of each model indicates that Swin outperforms the others, suggesting its significant potential as the best model with further adjustments and fine-tuning.

\begin{table}[h]
  \centering
  \caption{Recall metrics of Training 1 for ResNet50, ViT, and Swin}\label{tab:recall}
  \begin{tabular}{@{}lcccccccccc@{}}
    \toprule
    & \textbf{Resnet-50} & \textbf{ViT} & \textbf{Swin} \\
    \midrule
    \textbf{Steel Sheets} & 0.9667 & 1.0000 & 1.0000 \\
    \textbf{Stamping Scrap} & 0.9467 & 0.9800 & 0.9800 \\
    \textbf{Swarf Scrap} & 1.0000 & 1.0000 & 1.0000 \\
    \textbf{Low-Quality Oxyfuel Cutting} & 0.9091 & 0.8636 & 0.9545 \\
    \textbf{High-Quality Oxyfuel Cutting} & 0.9831 & 0.9661 & 0.9746 \\
    \textbf{Low-Quality Packages} & 0.9176 & 0.9765 & 0.9529 \\
    \textbf{High-Quality Packages} & 0.9250 & 0.8875 & 0.9250 \\
    \textbf{Shredder} & 1.0000 & 1.0000 & 1.0000 \\
    \textbf{Sheared Scrap} & 0.8475 & 0.9237 & 0.7966 \\
    \midrule
    \textbf{Model Accuracy} & 0.9392 & 0.9526 & 0.9465 \\
    \bottomrule
  \end{tabular}
\end{table}

\subsection{Experiment \#2: Conformal Prediction}

\begin{figure}[h]
  \centering
  \begin{minipage}{0.3\textwidth}
    \centering
    \includegraphics[width=\linewidth]{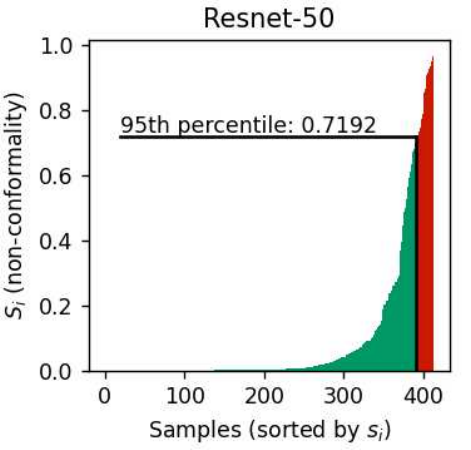}
  \end{minipage}
  \begin{minipage}{0.3\textwidth}
    \centering
    \includegraphics[width=\linewidth]{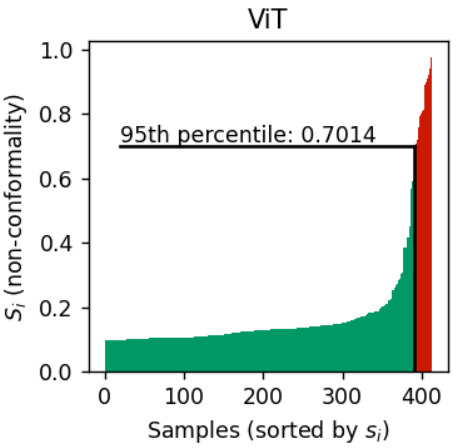}
  \end{minipage}
  \begin{minipage}{0.3\textwidth}
    \centering
    \includegraphics[width=\linewidth]{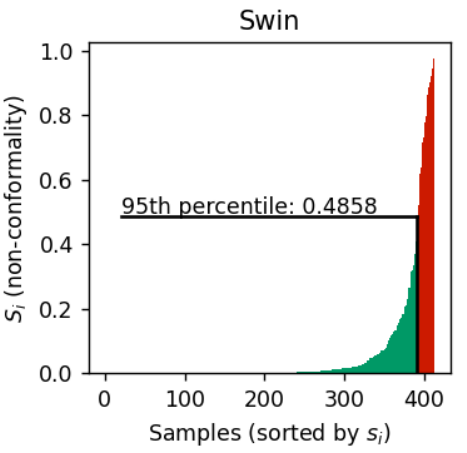}
  \end{minipage}
  
  \caption{Calibration Conformal Prediction Threshold for Resnet-50, ViT and Swin.}\label{fig:calibration_conformal}
\end{figure}

In this experiment, we employed the Split Conformal Prediction method to determine thresholds for assessing model uncertainties. Figure \ref{fig:calibration_conformal} presents the distribution of the $s_i$ scores for each sample in the calibration set, sorted in ascending order, along with the calibrated thresholds for each model. Notably, the Swin model exhibits the lowest threshold among the evaluated models. As detailed in Subsection \ref{sec_methods_conformal}, a lower threshold signifies that the Swin model demonstrates a higher level of confidence in its predictions for 95\% of the samples in the calibration dataset compared with both the ResNet-50 and ViT models.

Predictions for the test dataset were generated by employing the thresholds calibrated for each model. This results in a set of potential classes whose $s_i$ scores fall below the respective thresholds. Table \ref{tab:set_size_metrics} details the average size of the prediction sets for each class and model. An average set size of 1 signifies that the model predicted a single class for all inferences, indicating a higher certainty in the prediction. Conversely, an average set size greater than 1 suggests the inclusion of multiple classes within the predictions, whereas a set size lower than 1 implies instances where the predictions could not decisively identify any class.

\begin{table}[h]
  \centering
  \caption{Average Set Size Metrics for ResNet-50, ViT, and Swin Models}\label{tab:set_size_metrics}
  \begin{tabular}{@{}lccc@{}}
    \toprule
    & \textbf{ResNet-50} & \textbf{ViT} & \textbf{Swin} \\
    \midrule
    \textbf{Steel Sheets} & 1.0000 & 1.0000 & 1.0000 \\
    \textbf{Stamping Scrap} & 1.0400 & 1.0133 & 1.0000 \\
    \textbf{Swarf Scrap} & 1.0000 & 1.0000 & 1.0000 \\
    \textbf{Low-Quality Oxyfuel Cutting} & 1.1364 & 1.0455 & 0.9773 \\
    \textbf{High-Quality Oxyfuel Cutting} & 1.0169 & 1.0000 & 0.9661 \\
    \textbf{Low-Quality Packages} & 1.1190 & 1.0000 & 0.9762 \\
    \textbf{High-Quality Packages} & 1.0500 & 1.0250 & 0.9750 \\
    \textbf{Shredder} & 1.0000 & 1.0000 & 1.0000 \\
    \textbf{Sheared Scrap} & 1.2034 & 1.0508 & 1.0000 \\
    \midrule
    \textbf{Overall} & 1.0707 & 1.0171 & 0.9878 \\
    \bottomrule
  \end{tabular}
\end{table}

Assuming that outcomes with a set size of 1, which accurately identify a true class, represent correct predictions, it is feasible to calculate the coverage for each class. In the domain of Conformal Prediction, coverage functions similarly to recall, serving as an indicator of the proportion of correct predictions made. Table \ref{tab:coverage_metrics} delineates the coverage for each class and the overall coverage for each model, highlighting that the Vision Transformer (ViT) and Swin Transformer models exhibit a tie in overall coverage. However, one can examine which model yielded more uncertain predictions, defined as those with a set size diverging from 1. Specifically, the uncertain predictions were as follows: for ResNet-50, there were 28 predictions, each with a set size of 2; for ViT, there were 6 predictions, also with a set size of 2; and for Swin, there were 5 predictions, each with a set size of 0, indicating no class could be definitively assigned. This analysis reveals that the Swin model produced fewer uncertain predictions compared to the other models, suggesting its superior capability in generating more definitive and reliable classifications.

\begin{table}[h]
  \centering 
  \caption{Coverage Metrics for ResNet-50, ViT, and Swin Models}\label{tab:coverage_metrics}
  \begin{tabular}{@{}lccc@{}}
    \toprule
    & \textbf{ResNet-50} & \textbf{ViT} & \textbf{Swin} \\
    \midrule
    \textbf{Steel Sheets} & 1.0000 & 1.0000 & 1.0000 \\
    \textbf{Stamping Scrap} & 0.9333 & 0.9733 & 0.9867 \\
    \textbf{Swarf Scrap} & 1.0000 & 1.0000 & 1.0000 \\
    \textbf{Low-Quality Oxyfuel Cutting} & 0.8182 & 0.8182 & 0.9091 \\
    \textbf{High-Quality Oxyfuel Cutting} & 0.9661 & 0.9831 & 0.9661 \\
    \textbf{Low-Quality Packages} & 0.8571 & 0.9762 & 0.9524 \\
    \textbf{High-Quality Packages} & 0.9250 & 0.9250 & 0.9500 \\
    \textbf{Shredder} & 1.0000 & 1.0000 & 1.0000 \\
    \textbf{Sheared Scrap} & 0.7458 & 0.9153 & 0.8475 \\
    \midrule
    \textbf{Overall} & 0.9049 & 0.9512 & 0.9512 \\
    \bottomrule
  \end{tabular}
\end{table}

\subsection{Experiment \#3: Explainability}

\begin{figure}[H]%
\centering
\includegraphics[width=0.7\textwidth]{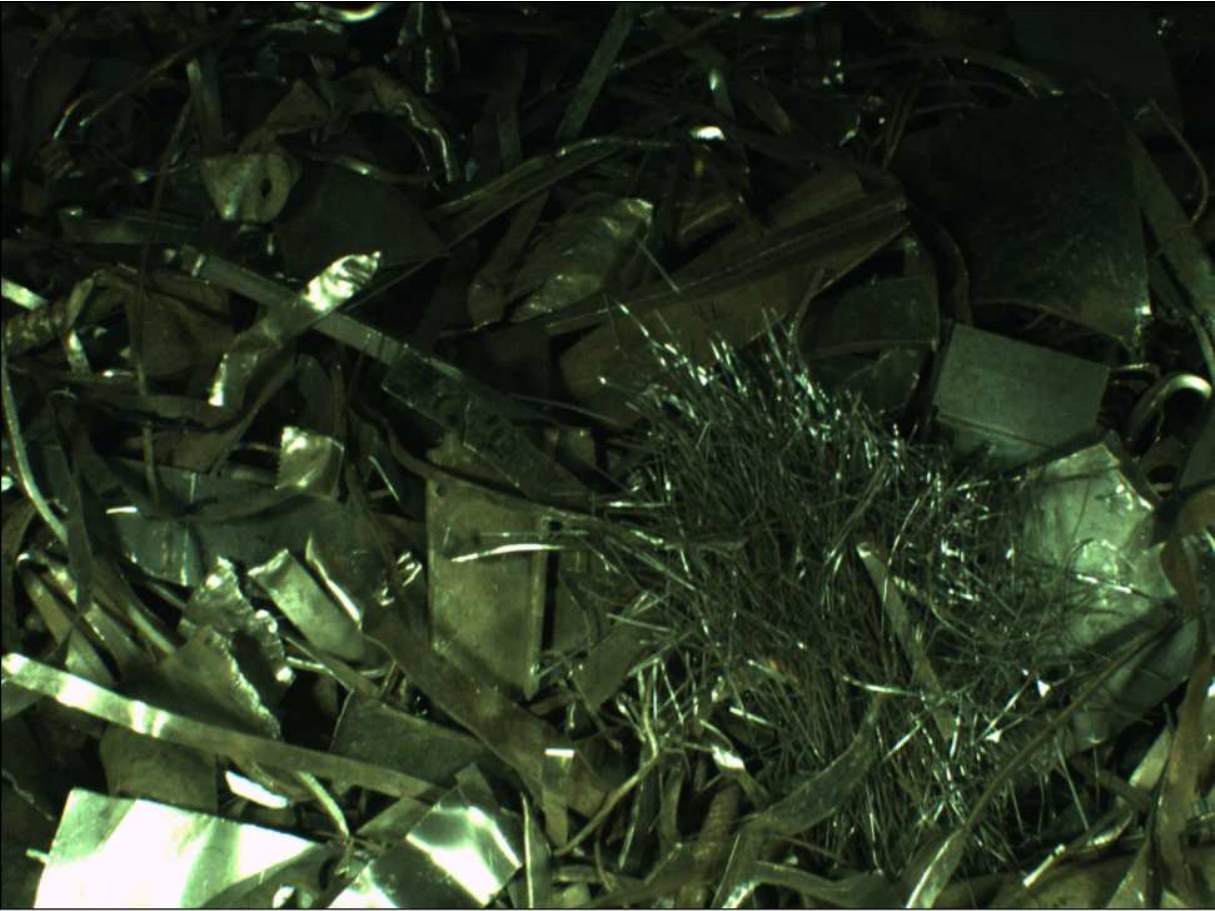}
\caption{Input Image for explainability methods comparison.}\label{fig:input_exp}
\end{figure}

For the analysis of the explainability methods applied to different models, specific dataset samples were carefully selected. Figure \ref{fig:input_exp} serves as a prime example of a challenging instance. This image, categorized under the High-Quality Oxyfuel Cutting class, presents a unique analytical opportunity because of the presence of a region that exhibits a pattern distinct from the class's standard characteristics. Such variance positions this sample as a strong case for evaluating the effectiveness of various explainability techniques in elucidating the nuances of model decision-making criteria.

\begin{figure}

  \centering
  \begin{minipage}{0.3\textwidth}
    \centering
    \includegraphics[width=\linewidth]{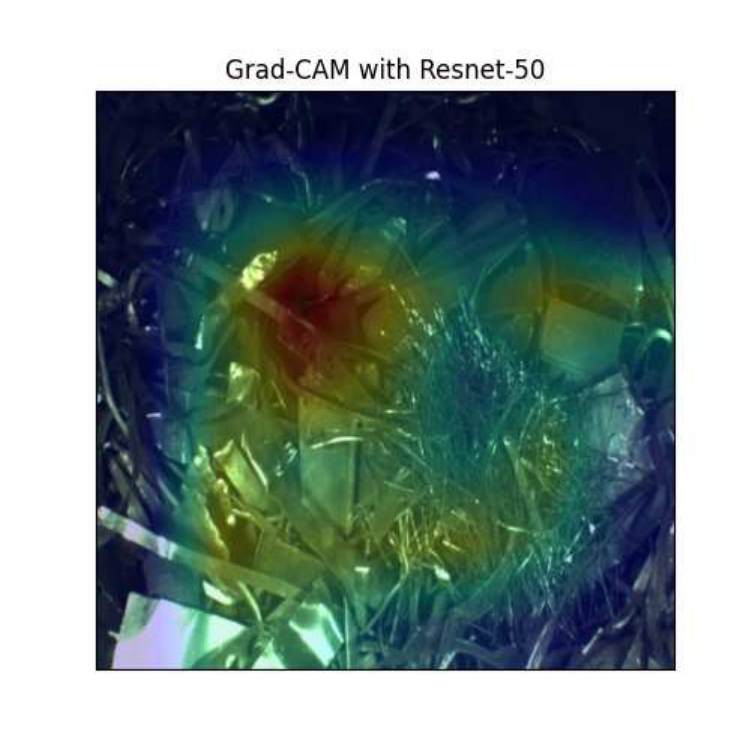}
  \end{minipage}
  \begin{minipage}{0.3\textwidth}
    \centering
    \includegraphics[width=\linewidth]{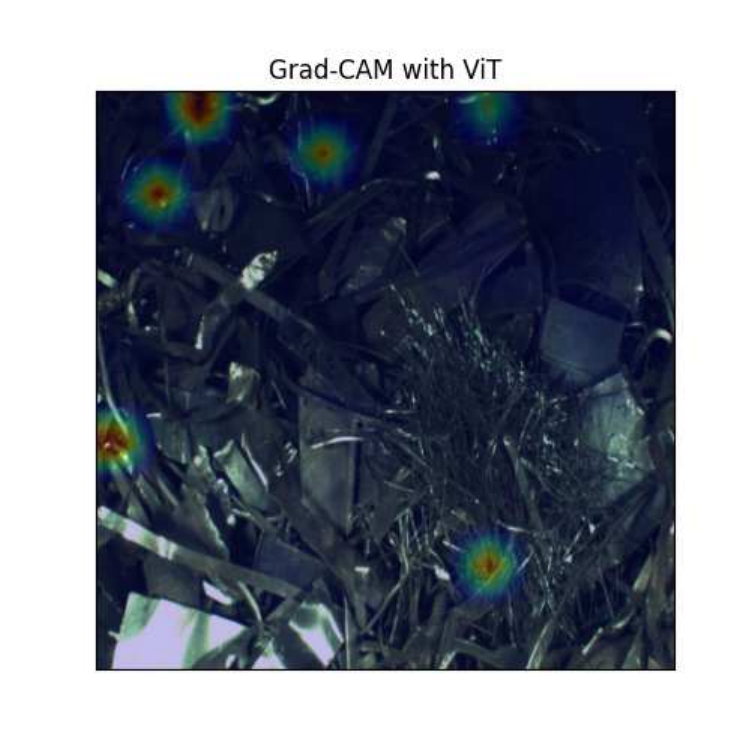}
  \end{minipage}
  \begin{minipage}{0.3\textwidth}
    \centering
    \includegraphics[width=\linewidth]{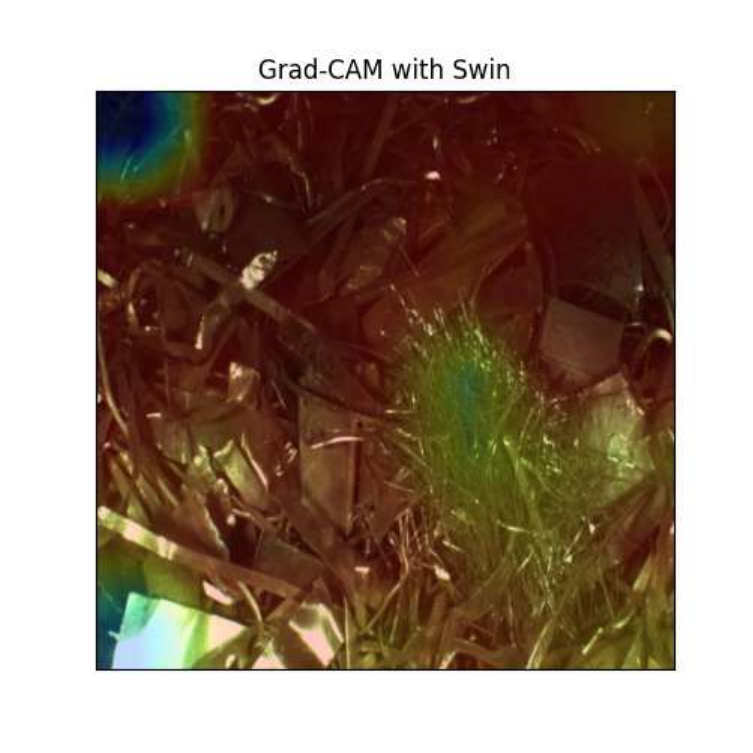}
  \end{minipage}
  
  \begin{minipage}{0.3\textwidth}
    \centering
    \includegraphics[width=\linewidth]{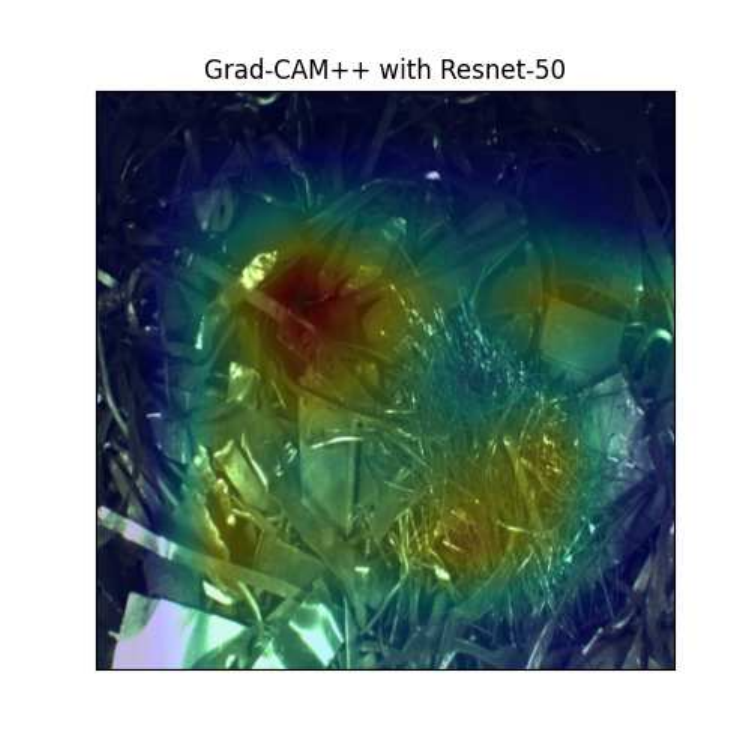}
  \end{minipage}
  \begin{minipage}{0.3\textwidth}
    \centering
    \includegraphics[width=\linewidth]{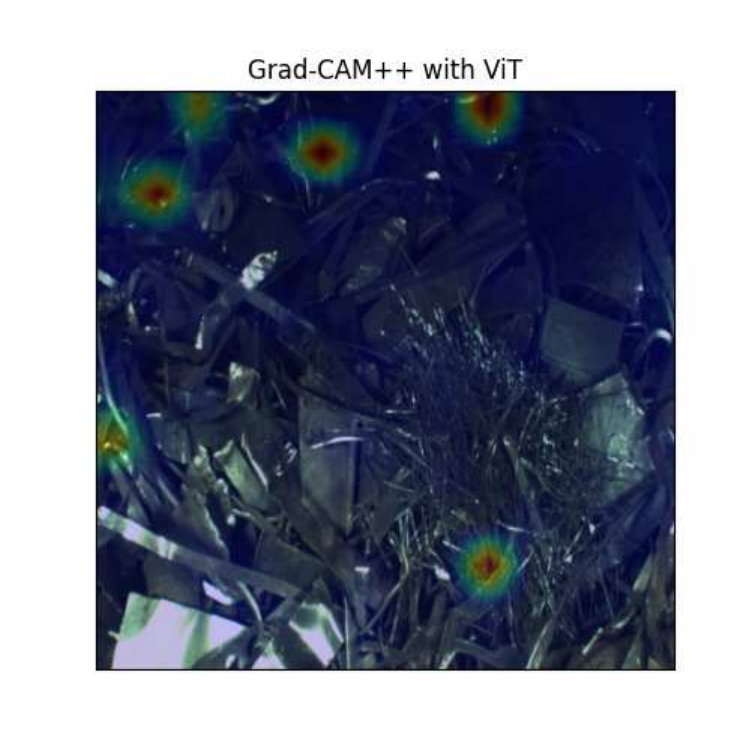}
  \end{minipage}
  \begin{minipage}{0.3\textwidth}
    \centering
    \includegraphics[width=\linewidth]{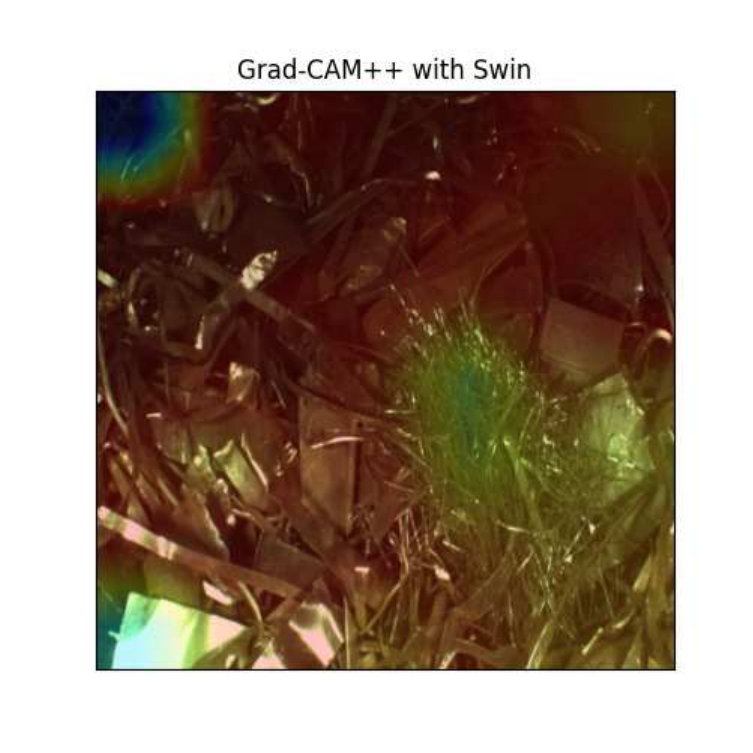}
  \end{minipage}
  
  \begin{minipage}{0.3\textwidth}
    \centering
    \includegraphics[width=\linewidth]{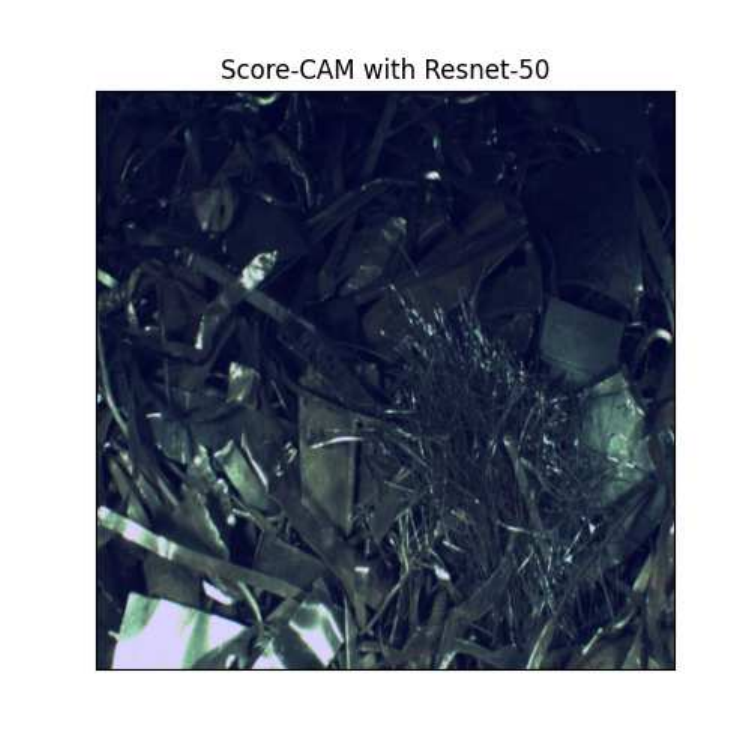}
  \end{minipage}
  \begin{minipage}{0.3\textwidth}
    \centering
    \includegraphics[width=\linewidth]{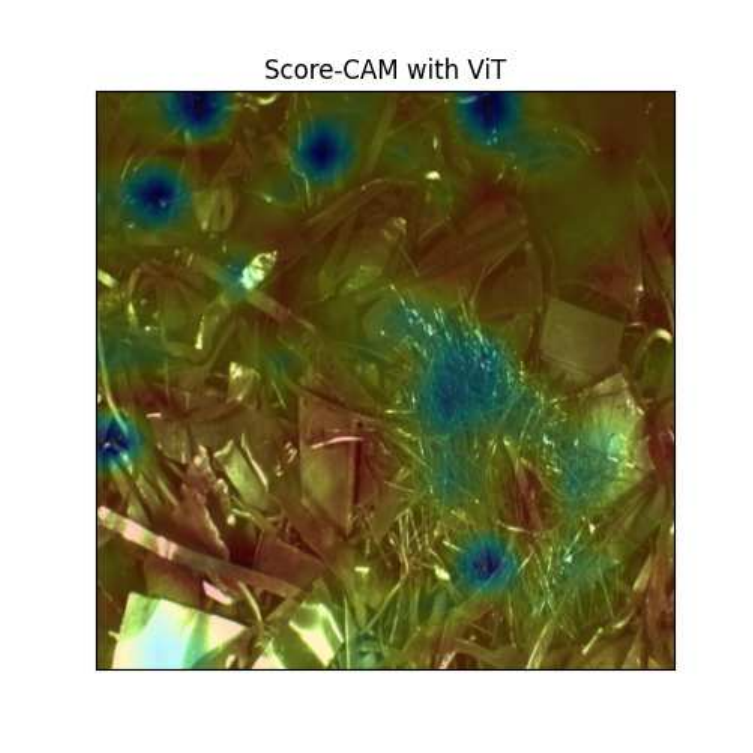}
  \end{minipage}
  \begin{minipage}{0.3\textwidth}
    \centering
    \includegraphics[width=\linewidth]{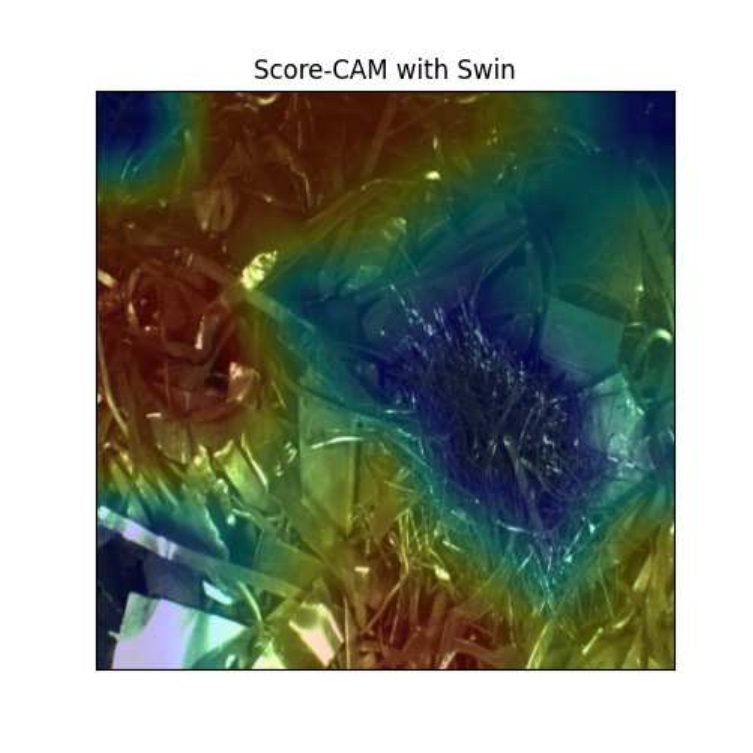}
  \end{minipage}

    \begin{minipage}{0.3\textwidth}
    \centering
    \includegraphics[width=\linewidth]{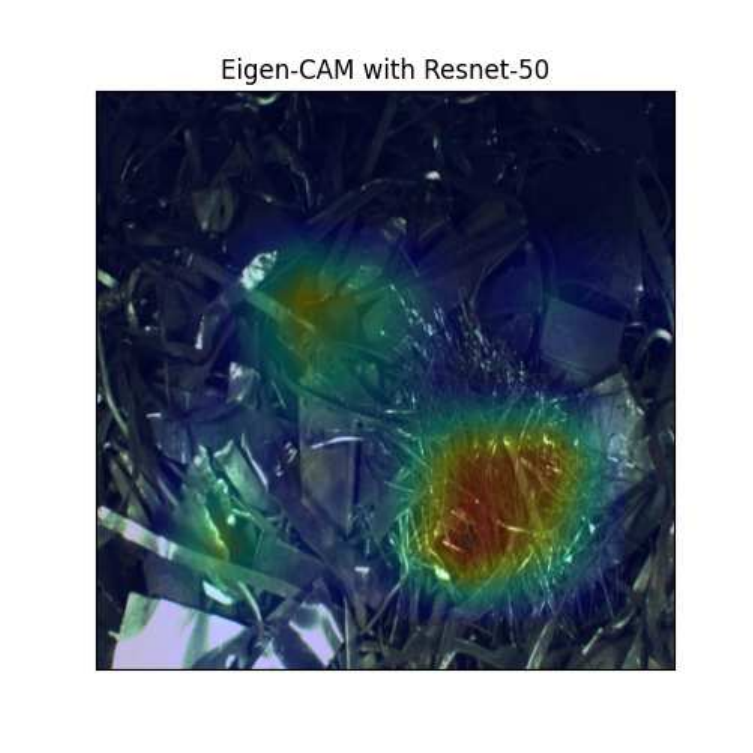}
  \end{minipage}
  \begin{minipage}{0.3\textwidth}
    \centering
    \includegraphics[width=\linewidth]{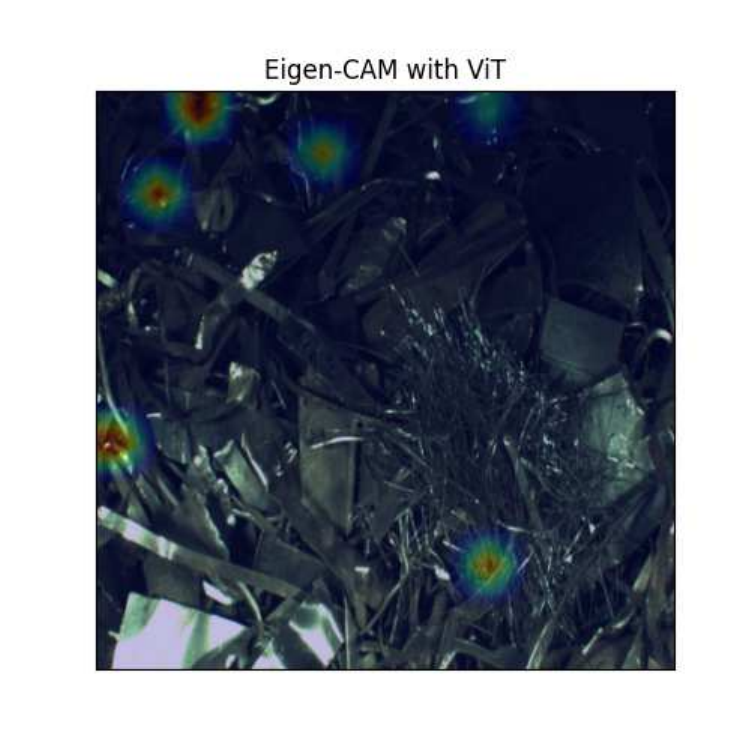}
  \end{minipage}
  \begin{minipage}{0.3\textwidth}
    \centering
    \includegraphics[width=\linewidth]{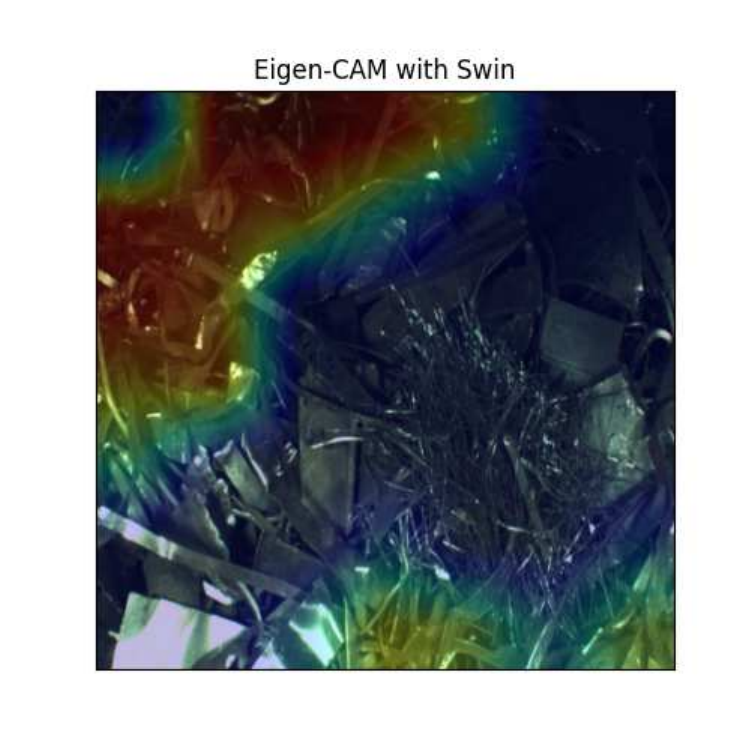}
  \end{minipage}
  
  \caption{CAM Methods result for each model.}\label{fig:cam_results}
\end{figure}

For each model, a variety of target layers were explored to optimize the application of explainability methods. Figure \ref{fig:cam_results} showcases the outcomes for CAM-based methods, identifying the most effective target layer for each model. These methods were specifically applied to activation maps pertinent to the High-Quality Oxyfuel Cutting classification.

In the case of the ResNet-50 model, CAM-based explainability techniques proved to be less effective. Among these, Grad-CAM and Grad-CAM++ emerged as the more favorable options; however, they still did not yield reliably interpretable results. For the Vision Transformer (ViT) model, Score-CAM was identified as the superior method, albeit with limited accuracy. Conversely, the methods demonstrated superior performance with the Swin model, where Score-CAM again stood out as the most effective. As illustrated in Figure \ref{fig:cam_results}, the application of Score-CAM to the Swin model reveals that the region diverging in pattern from the typical High-Quality Oxyfuel Cutting characteristics is not highlighted as significantly relevant in the heat map. This observation underscores the nuanced capability of Score-CAM with the Swin model to discern and emphasize areas crucial for accurate classification.

\begin{figure}[H]

  \centering
  \begin{minipage}{0.5\textwidth}
  \centering
    Resnet-50\\
    \centering
    \includegraphics[width=\linewidth]{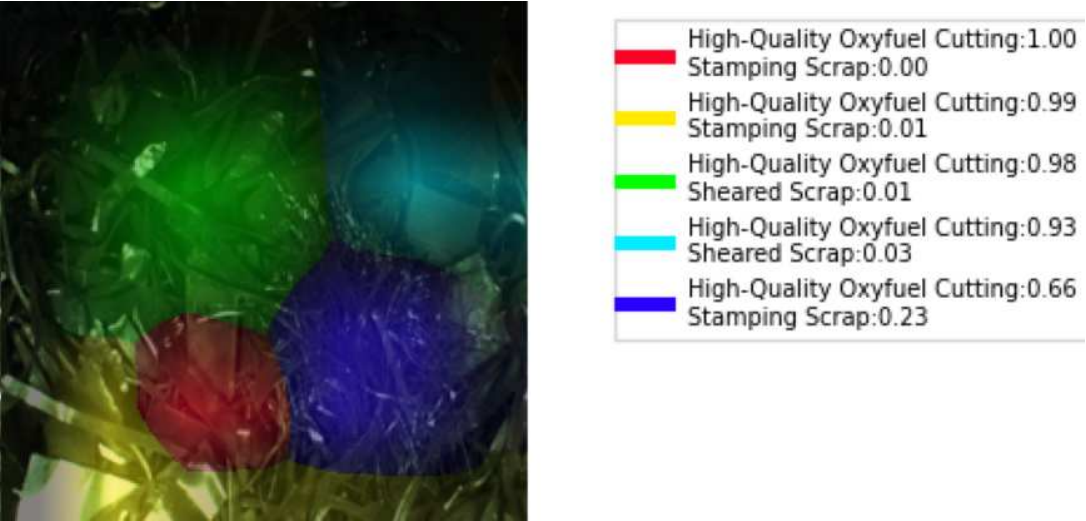}
  \end{minipage}
  
  \bigskip
  
  \begin{minipage}{0.5\textwidth}
  \centering
    ViT\\
    \centering
    \includegraphics[width=\linewidth]{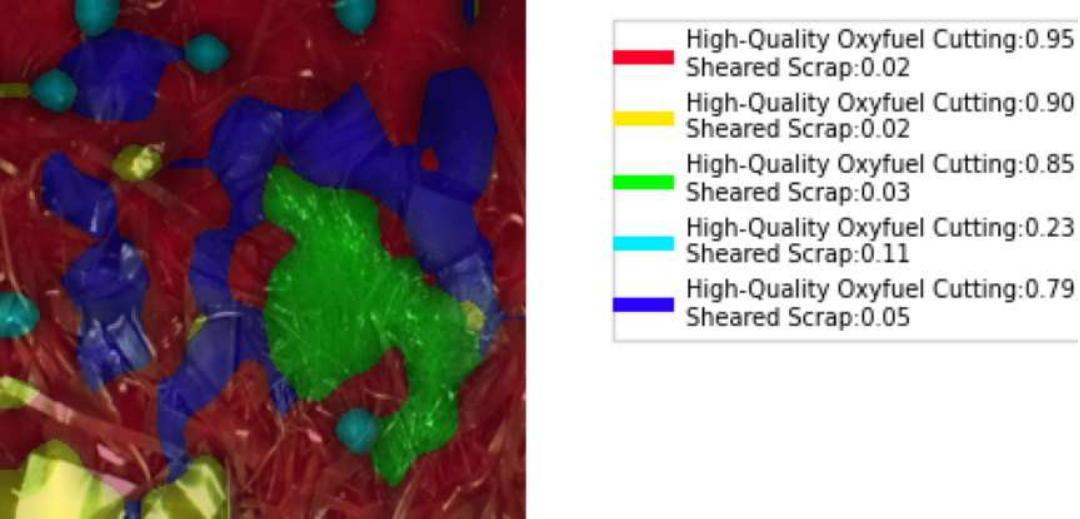}
  \end{minipage}
  
  \bigskip
  
  \begin{minipage}{0.5\textwidth}
    \centering
    Swin\\
    \centering
    \includegraphics[width=\linewidth]{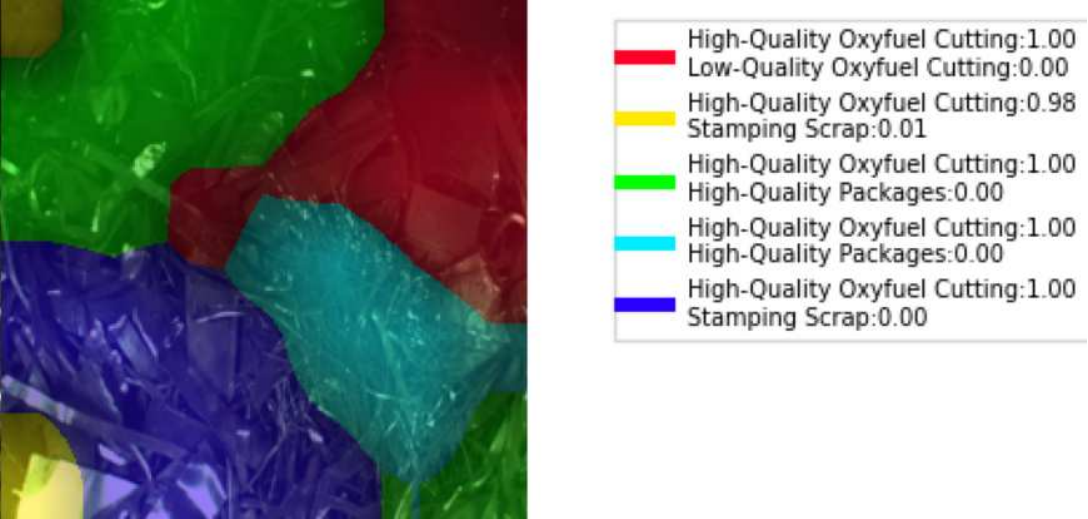}
  \end{minipage}
  
  \caption{Deep Feature Factorization (DFF) results for each model.}\label{fig:dff_results}
\end{figure}

The Deep Feature Factorization technique segments the input image in such a manner that the region deviating in pattern from the High-Quality Oxyfuel Cutting norm is distinctly highlighted. Notably, it was solely within the ResNet-50 model that this method identified a region exhibiting lower confidence in the High-Quality Oxyfuel Cutting classification, as depicted in Figure \ref{fig:dff_results}. This specificity underscores the nuanced capacity of the Deep Feature Factorization approach to discern and evaluate segments within an image based on their conformity to class characteristics.

Consequently, the combination of Score-CAM with the Swin model emerged as the most effective method for elucidating the explainability aspects of scrap classification. In an effort to decipher the predominant features leveraged by the model for scrap categorization and to understand the underlying causes of class confusion, Score-CAM was deployed in various scenarios, detailed subsequently.

\begin{figure}[h]
\centering
\includegraphics[width=0.8\textwidth]{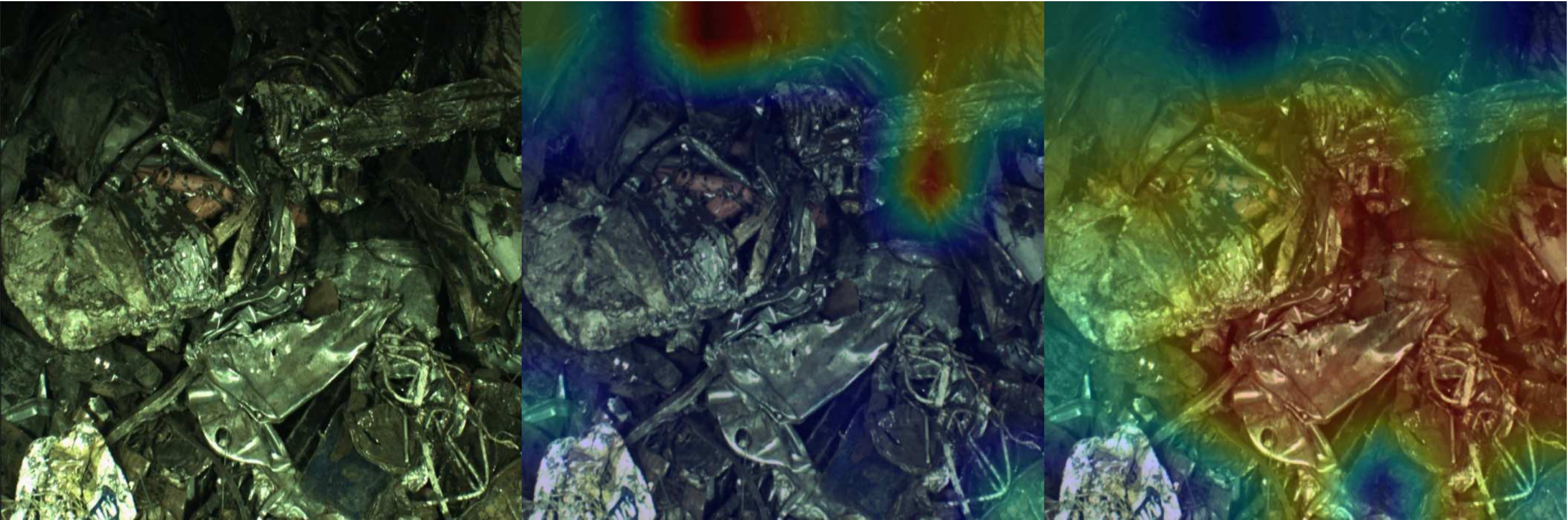}

    \begin{minipage}{0.2\textwidth}
        \centering
        (a)
    \end{minipage}
    \begin{minipage}{0.2\textwidth}
        \centering
        (b)
    \end{minipage}
    \begin{minipage}{0.2\textwidth}
        \centering
        (c)
    \end{minipage}
\caption{Understanding Sheared Scrap classification with Score-CAM for Swin model, \textbf{(a)} is the input, \textbf{(b)} is the heat map for Low-Quality Oxyfuel Cutting Scrap class, and \textbf{(c)} is the heat map for Sheared Scrap. }\label{fig:sheared_lq-oxy_vs_sheared}
\end{figure}

The classes most commonly confused by the models are Low-Quality Oxyfuel Cutting Scrap and Sheared Scrap, and vice versa. This confusion arises because, in certain instances, these classes bear striking similarities, sharing the same raw material but differing in the cutting process employed. Figure \ref{fig:sheared_lq-oxy_vs_sheared}b displays a heat map for Low-Quality Oxyfuel Cutting Scrap and Figure \ref{fig:sheared_lq-oxy_vs_sheared}c a heat map for Sheared Scrap. The depicted input image was categorized as Sheared Scrap, with the model attributing significant weight to the crumpled aspect of the scrap to determine its classification as Sheared Scrap. This classification was made with a confidence level of 0.82 and a $s_i$ score of 0.18. The $s_i$ score falls below the conformal prediction threshold for the Swin model, which stands at 0.4858, indicating high reliability in this particular prediction.

\begin{figure}[h]%
\centering
\includegraphics[width=0.8\textwidth]{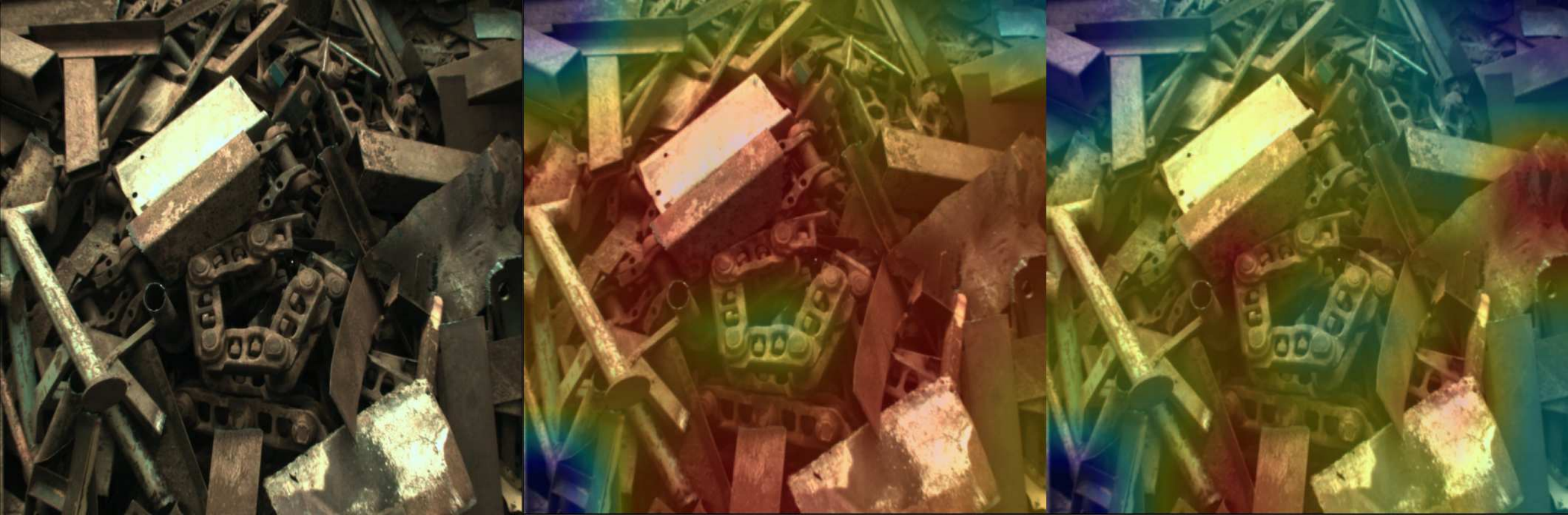}

    \begin{minipage}{0.2\textwidth}
        \centering
        (a)
    \end{minipage}
    \begin{minipage}{0.2\textwidth}
        \centering
        (b)
    \end{minipage}
    \begin{minipage}{0.2\textwidth}
        \centering
        (c)
    \end{minipage}
\caption{Understanding Low-Quality Oxyfuel Cutting Scrap classification with Score-CAM for Swin model, \textbf{(a)} is the input, \textbf{(b)} is the heat map for Low-Quality Oxyfuel Cutting class, and \textbf{(c)} is the heat map for Sheared Scrap. }\label{fig:lq-oxy-lq-oxy_vs_sheared}
\end{figure}

Figure \ref{fig:lq-oxy-lq-oxy_vs_sheared}a presents an instance of Low-Quality Oxyfuel Cutting Scrap, with Figure \ref{fig:lq-oxy-lq-oxy_vs_sheared}b highlighting the features deemed most indicative of this class, and Figure \ref{fig:lq-oxy-lq-oxy_vs_sheared}c emphasizing those associated with Sheared Scrap. This example was accurately identified by the model with a high confidence of 0.99 and an $s_i$ score of 0.1, both figures lying comfortably below the established conformal threshold. Observations from the heat map reveal a more pronounced signature for Low-Quality Oxyfuel Cutting Scrap compared with that of Sheared Scrap. Nevertheless, the heat map for Sheared Scrap delineates numerous areas that the model recognizes as potentially characteristic of Sheared Scrap, demonstrating the model's ability to discern and attribute relevance to various regions within the image.

\begin{figure}[h]%
\centering
\includegraphics[width=0.8\textwidth]{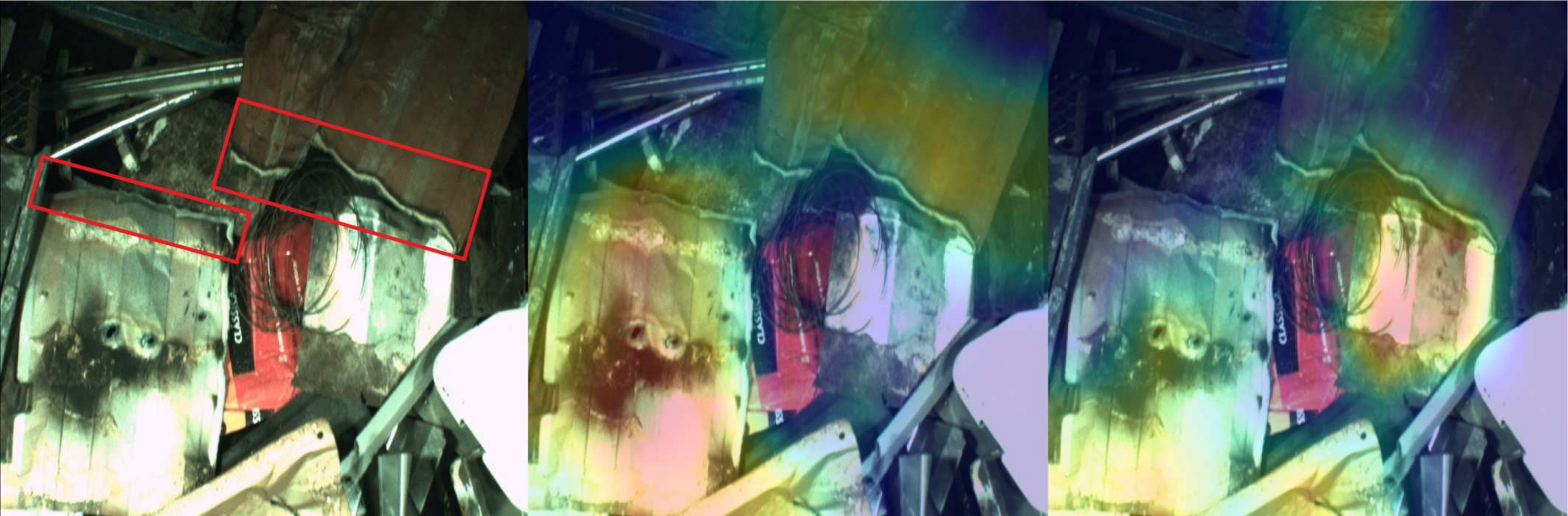}

    \begin{minipage}{0.2\textwidth}
        \centering
        (a)
    \end{minipage}
    \begin{minipage}{0.2\textwidth}
        \centering
        (b)
    \end{minipage}
    \begin{minipage}{0.2\textwidth}
        \centering
        (c)
    \end{minipage}
\caption{Understanding why the model confuses the Sheared Scrap classification with Score-CAM for Swin model, \textbf{(a)} is the input, \textbf{(b)} is the heat map for Low-Quality Oxyfuel Cutting class, and \textbf{(c)} is the heat map for Sheared Scrap. }\label{fig:sheared_as_oxy-lq_with_indication}
\end{figure}

In Figure \ref{fig:sheared_as_oxy-lq_with_indication}a, the input image is initially labeled as Sheared Scrap. Notably, the red boxes highlight the presence of burning marks on the scrap, which is indicative of the oxyfuel cutting process. Despite its original classification, this image was incorrectly categorized as Low-Quality Oxyfuel Cutting Scrap, evidenced by an $s_i$ score of 0.14. Intriguingly, the heat map in Figure \ref{fig:sheared_as_oxy-lq_with_indication}b, corresponding to Low-Quality Oxyfuel Cutting Scrap, accentuates the scrap sections bearing burning marks, with a notably more extensive hot area compared to the heat map for Sheared Scrap in Figure \ref{fig:sheared_as_oxy-lq_with_indication}c. This observation suggests that burning marks are a significant feature for the classification of Low-Quality Oxyfuel Cutting Scrap, highlighting the model's sensitivity to specific visual cues associated with the cutting process.


\begin{figure}[h]%
\centering
\includegraphics[width=0.8\textwidth]{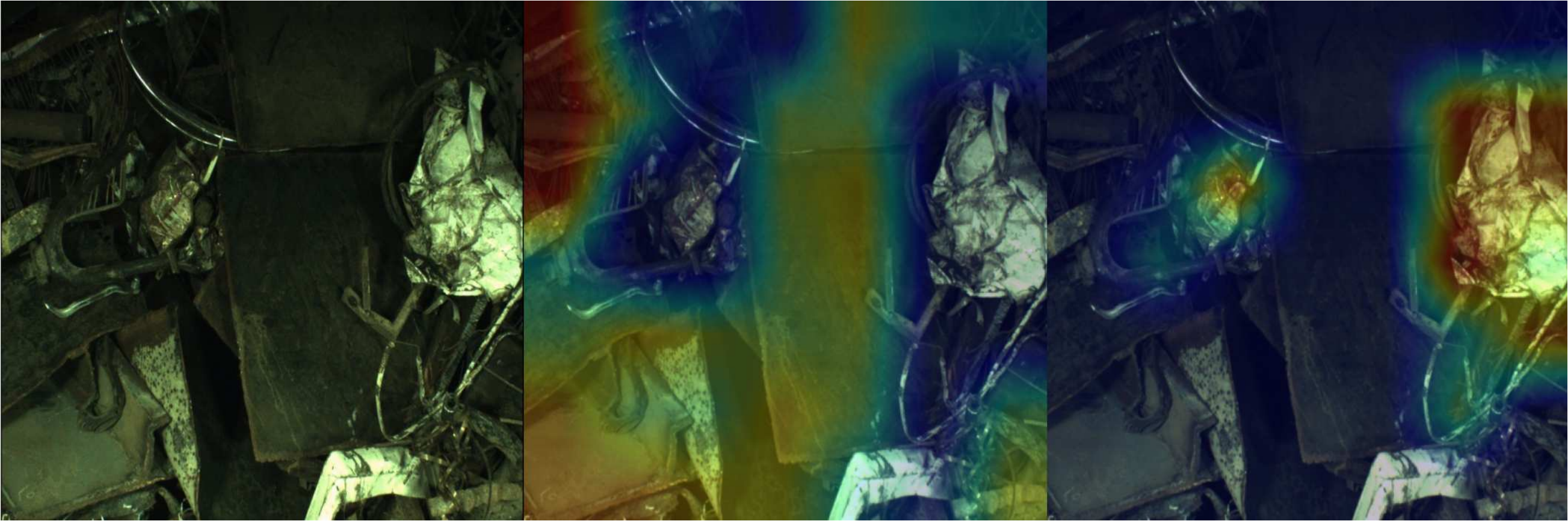}

    \begin{minipage}{0.2\textwidth}
        \centering
        (a)
    \end{minipage}
    \begin{minipage}{0.2\textwidth}
        \centering
        (b)
    \end{minipage}
    \begin{minipage}{0.2\textwidth}
        \centering
        (c)
    \end{minipage}
\caption{Understanding why the model confuses the Low-Quality Oxyfuel Cutting Scrap classification with Score-CAM for Swin model, \textbf{(a)} is the input, \textbf{(b)} is the heat map for Low-Quality Oxyfuel Cutting class, and \textbf{(c)} is the heat map for Sheared Scrap. }\label{fig:oxy-lq_as_sheared.pdf}
\end{figure}

An additional illustration of the significance of burning and cutting marks is provided in Figure \ref{fig:oxy-lq_as_sheared.pdf}. Here, the input, originally labeled as Low-Quality Oxyfuel Cutting Scrap, was categorized as Sheared Scrap, yielding an $s_i$ score of 0.42. This score falls below the conformal prediction threshold, albeit narrowly. Observations from Figure \ref{fig:oxy-lq_as_sheared.pdf}b reveal that the heat map for Low-Quality Oxyfuel Cutting Scrap highlights regions marked by oxyfuel cutting, while prominently bypassing crumpled areas. Conversely, the heat map for Sheared Scrap in Figure \ref{fig:oxy-lq_as_sheared.pdf}c demonstrates the model's prioritization of crumpled areas, suggesting that such features hold considerable weight in the classification of Sheared Scrap. Notably, the crumpled regions within the Low-Quality Oxyfuel Cutting Scrap were deemed more significant than the oxyfuel cutting marks, leading the model to classify this sample as Sheared Scrap, with an $s_i$ score approaching the threshold. This insight underscores the model's interpretation of visual cues, differentiating between similar classes based on distinctive textural characteristics.


\begin{figure}[h]%
\centering
\includegraphics[width=0.8\textwidth]{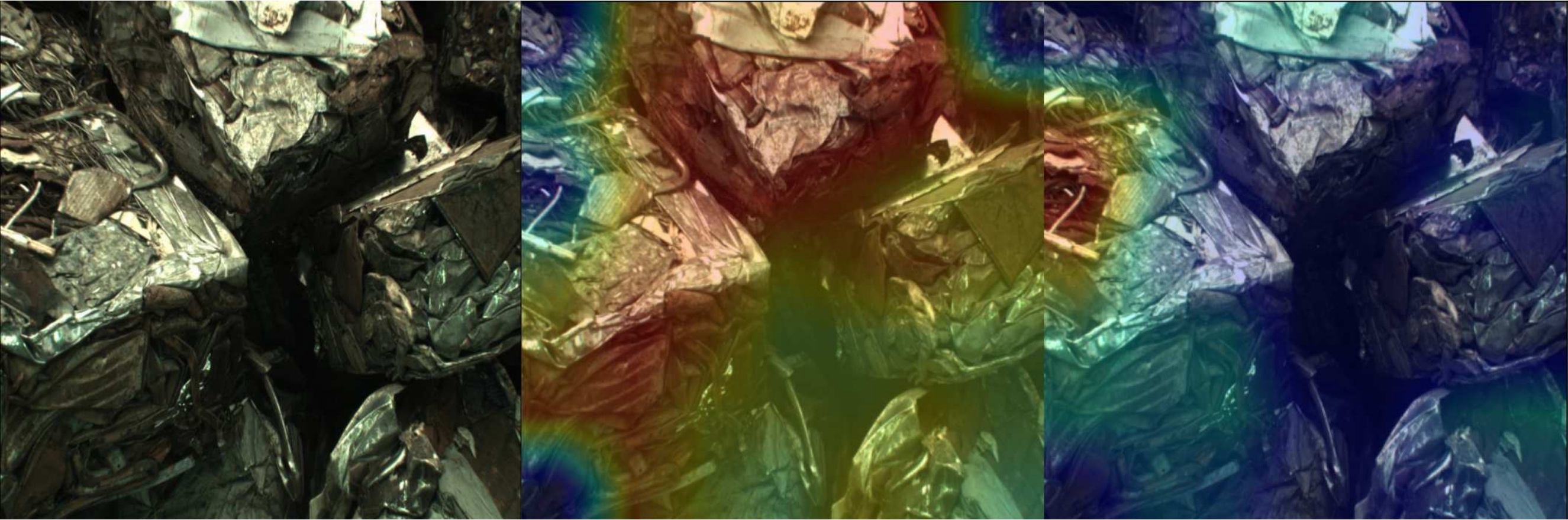}

    \begin{minipage}{0.2\textwidth}
        \centering
        (a)
    \end{minipage}
    \begin{minipage}{0.2\textwidth}
        \centering
        (b)
    \end{minipage}
    \begin{minipage}{0.2\textwidth}
        \centering
        (c)
    \end{minipage}
\caption{Understanding Low-Quality Packages classification with Score-CAM for Swin model, \textbf{(a)} is the input, \textbf{(b)} is the heat map for Low-Quality Packages class, and \textbf{(c)} is the heat map for Sheared Scrap. }\label{fig:lq-packages_it_vs_sheared}
\end{figure}

\begin{figure}[h]%
\centering
\includegraphics[width=0.8\textwidth]{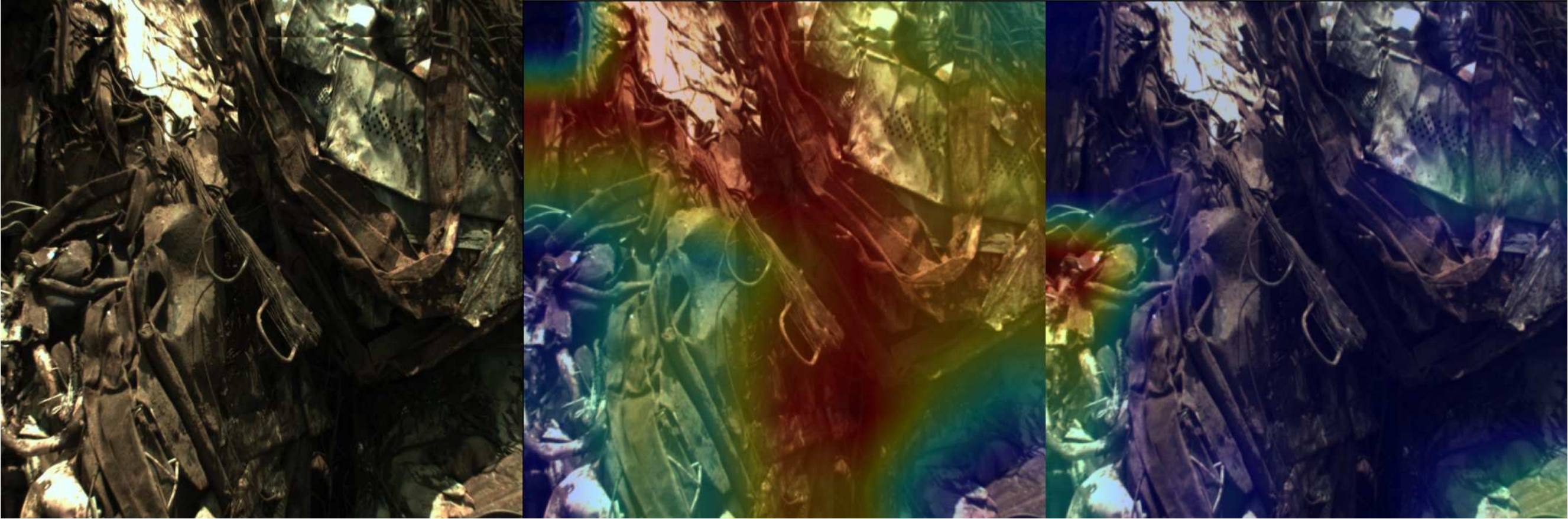}

    \begin{minipage}{0.2\textwidth}
        \centering
        (a)
    \end{minipage}
    \begin{minipage}{0.2\textwidth}
        \centering
        (b)
    \end{minipage}
    \begin{minipage}{0.2\textwidth}
        \centering
        (c)
    \end{minipage}
\caption{Understanding Low-Quality Packages classification with Score-CAM for Swin model, \textbf{(a)} is the input, \textbf{(b)} is the heat map for Low-Quality Packages class, and \textbf{(c)} is the heat map for Sheared Scrap. }\label{fig:lq-packages_it_vs_sheared_2}
\end{figure}

The composition of Low-Quality Packages closely mirrors that of Sheared Scrap, as evidenced by Figures \ref{fig:lq-packages_it_vs_sheared} and \ref{fig:lq-packages_it_vs_sheared_2}. These figures showcase instances of Low-Quality Packages being accurately classified, each with an $s_i$ score of 0.1. The heat maps in Figures \ref{fig:lq-packages_it_vs_sheared}b and \ref{fig:lq-packages_it_vs_sheared_2}b highlight the cube formation and intervening spaces as the most critical features for this classification. Despite the identical material composition of these classes, Figures \ref{fig:lq-packages_it_vs_sheared}c and \ref{fig:lq-packages_it_vs_sheared_2} illustrate the model's inability to detect features characteristic of Sheared Scrap in these samples, underscoring the distinct classification criteria applied by the model.


\begin{figure}[h]%
\centering
\includegraphics[width=0.8\textwidth]{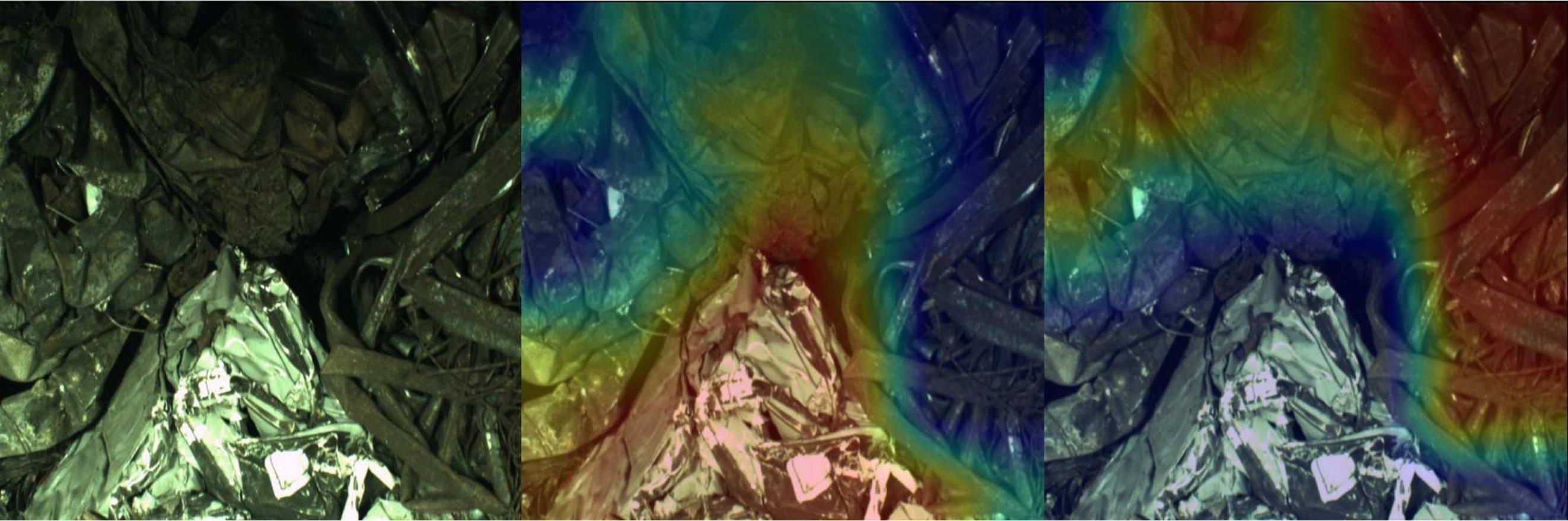}

    \begin{minipage}{0.2\textwidth}
        \centering
        (a)
    \end{minipage}
    \begin{minipage}{0.2\textwidth}
        \centering
        (b)
    \end{minipage}
    \begin{minipage}{0.2\textwidth}
        \centering
        (c)
    \end{minipage}
\caption{Understanding why the model confuses the Low-Quality Packages classification with Score-CAM for Swin model, \textbf{(a)} is the input, \textbf{(b)} is the heat map for Low-Quality Packages class, and \textbf{(c)} is the heat map for Sheared Scrap. }\label{fig:lq-packages_as_sheared}
\end{figure}

The singular occurrence of the model incorrectly classifying Low-Quality Packages as Sheared Scrap is illustrated in Figure \ref{fig:lq-packages_as_sheared}. In this case, the model erroneously identified the input as Sheared Scrap, assigning a $s_i$ score of 0.1. The heat map for Low-Quality Packages, visible in Figure \ref{fig:lq-packages_as_sheared}b, indicates that the model focused on an area featuring a single cube. However, due to the proximity of the other cubes, it appears that the model was unable to recognize the adjacent cubes as distinct entities. Conversely, Figure \ref{fig:lq-packages_as_sheared}c representing the Sheared Scrap heat map reveals that the model interpreted the surrounding regions as Sheared Scrap, attributing greater significance to these areas than to the lone cube for its classification decision. Thus, it becomes evident that when cubes within an image are not clearly delineated, the model encounters challenges in differentiating between Low-Quality Packages and Sheared Scrap, impacting its classification accuracy.


\begin{figure}[h]%
\centering
\includegraphics[width=0.8\textwidth]{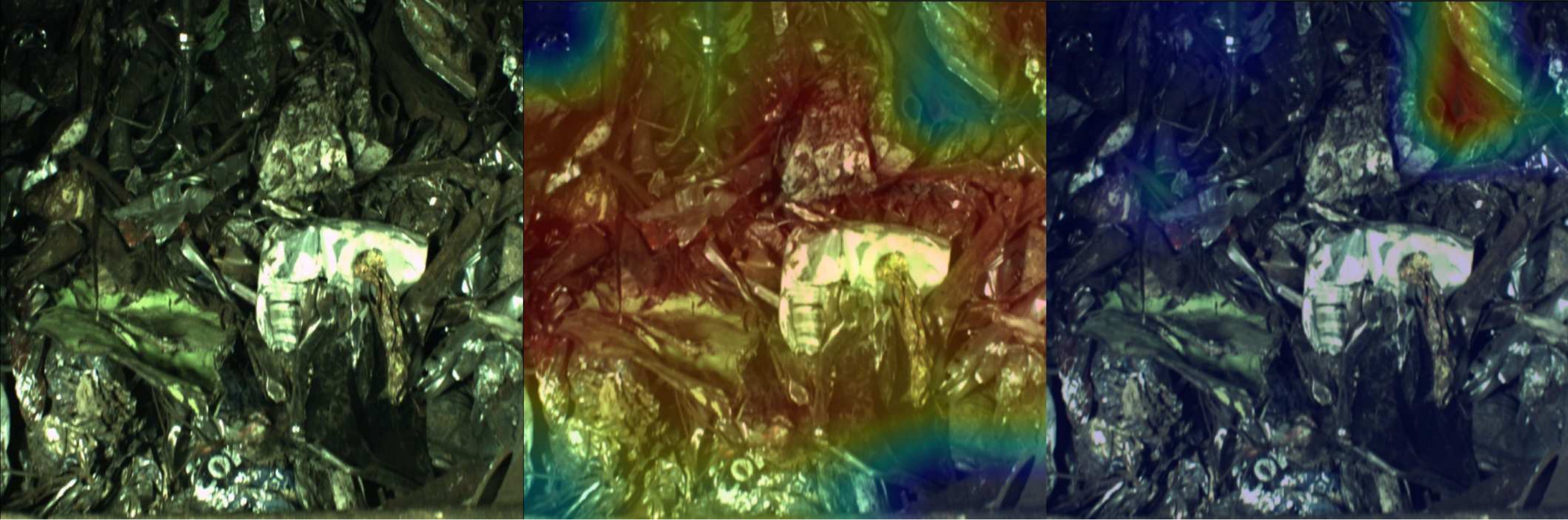}

    \begin{minipage}{0.2\textwidth}
        \centering
        (a)
    \end{minipage}
    \begin{minipage}{0.2\textwidth}
        \centering
        (b)
    \end{minipage}
    \begin{minipage}{0.2\textwidth}
        \centering
        (c)
    \end{minipage}
\caption{Understanding why the model confuses the Sheared Scrap classification with Score-CAM for Swin model, \textbf{(a)} is the input, \textbf{(b)} is the heat map for Low-Quality Packages class, and \textbf{(c)} is the heat map for Sheared Scrap. }\label{fig:sheared_as_lq-packages}
\end{figure}

\begin{figure}[h]%
\centering
\includegraphics[width=0.8\textwidth]{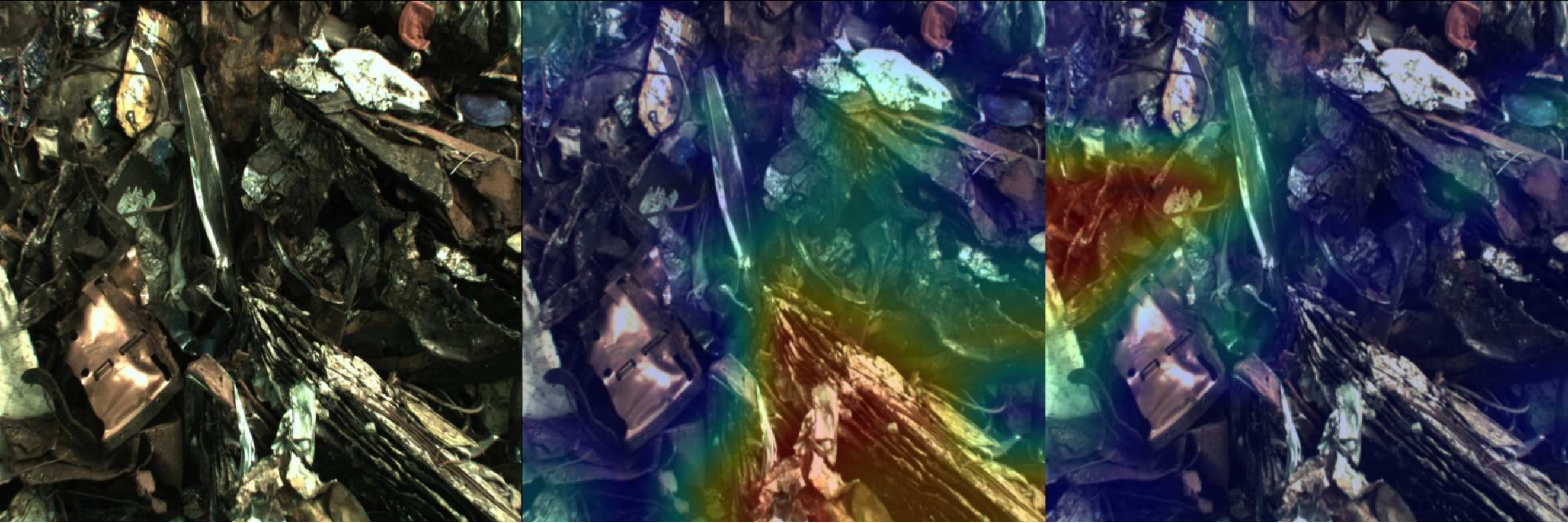}

    \begin{minipage}{0.2\textwidth}
        \centering
        (a)
    \end{minipage}
    \begin{minipage}{0.2\textwidth}
        \centering
        (b)
    \end{minipage}
    \begin{minipage}{0.2\textwidth}
        \centering
        (c)
    \end{minipage}
\caption{Understanding why the model confuses the Sheared Scrap classification with Score-CAM for Swin model, \textbf{(a)} is the input, \textbf{(b)} is the heat map for Low-Quality Packages class, and \textbf{(c)} is the heat map for Sheared Scrap. }\label{fig:sheared_as_lq-packages_2}
\end{figure}

When examining instances of misclassification between Sheared Scrap and Low-Quality Packages, as illustrated in Figures \ref{fig:sheared_as_lq-packages} and \ref{fig:sheared_as_lq-packages_2}, it becomes clear that the model erroneously identifies cube-like patterns within the scrap. These perceived cube formations in Figures \ref{fig:sheared_as_lq-packages}b and \ref{fig:sheared_as_lq-packages_2}b are mistakenly leveraged as the principal basis for classifying the samples as Low-Quality Packages, despite their true classification as Sheared Scrap. Specifically, for Figure \ref{fig:sheared_as_lq-packages}, the model generated an $s_i$ score of 0.06, and for Figure \ref{fig:sheared_as_lq-packages_2}, an $s_i$ score of 0.12. The corresponding heat maps for Sheared Scrap in Figures \ref{fig:sheared_as_lq-packages}c and \ref{fig:sheared_as_lq-packages_2}c fail to highlight significant regions, suggesting that these particular samples deviate from the typical pattern associated with Sheared Scrap. This discrepancy indicates a potential challenge for the model in recognizing and accurately classifying samples that diverge from the expected distribution patterns of Sheared Scrap.


\begin{figure}[h]%
\centering
\includegraphics[width=0.8\textwidth]{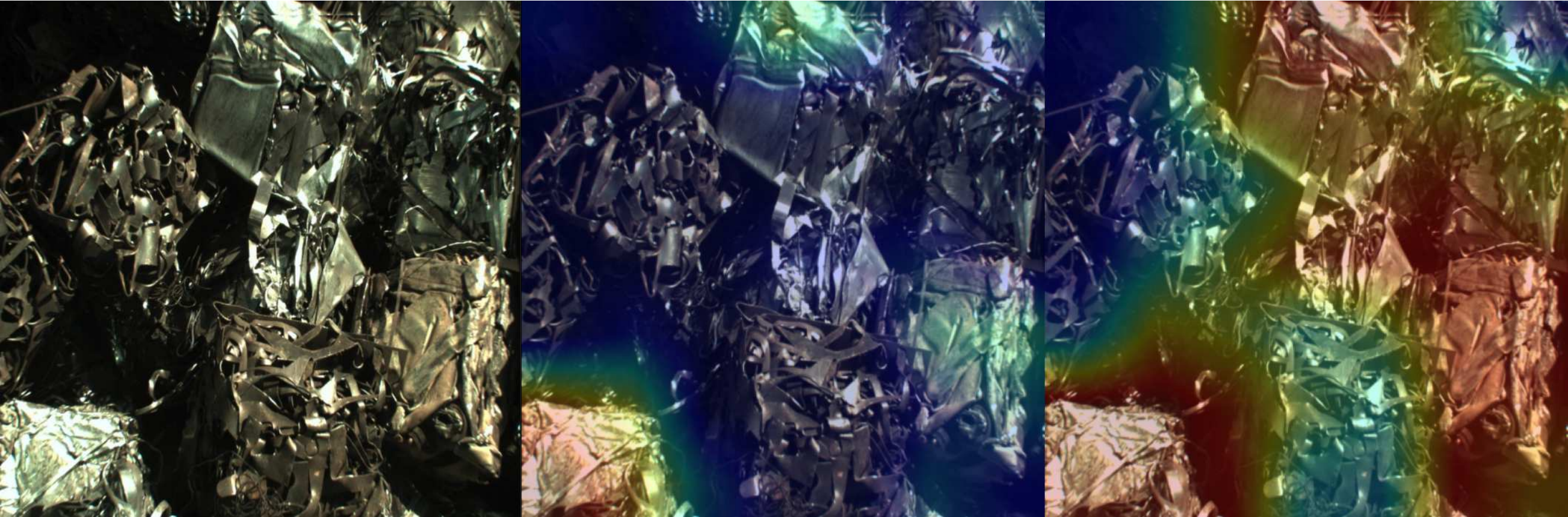}

    \begin{minipage}{0.2\textwidth}
        \centering
        (a)
    \end{minipage}
    \begin{minipage}{0.2\textwidth}
        \centering
        (b)
    \end{minipage}
    \begin{minipage}{0.2\textwidth}
        \centering
        (c)
    \end{minipage}
\caption{Understanding High-Quality Packages classification with Score-CAM for Swin model, \textbf{(a)} is the input, \textbf{(b)} is the heat map for Low-Quality Packages class, and \textbf{(c)} is the heat map for High-Quality Packages. }\label{fig:hq-packages_it_vs_lq-packages}
\end{figure}

It is also intriguing to explore how the model differentiates between Low-Quality Packages and High-Quality Packages. Figure \ref{fig:hq-packages_it_vs_lq-packages} showcases a correctly classified sample of High-Quality Packages, evidenced by an $s_i$ score of 0.001. The figure features two heat maps: Figure \ref{fig:hq-packages_it_vs_lq-packages}b for Low-Quality Packages and Figure \ref{fig:hq-packages_it_vs_lq-packages}c for High-Quality Packages. Despite the presence of numerous cubes in the image, the model adeptly identifies the cubes characteristic of High-Quality Packages. Notably, the heat map for Low-Quality Packages displays a significantly smaller area of interest, indicating the model's precision in distinguishing between the two package types based on subtle visual cues. This specificity highlights the model's capability to discern the higher organizational structure and clarity typical of High-Quality Packages.


The sample depicted in Figure \ref{fig:hq-packages_it_vs_lq-packages} consists of cubes that share similarities with both Stamping and High-Quality Oxyfuel Cutting. Interestingly, the heat map for High-Quality Packages, as illustrated in Figure \ref{fig:hq-packages_it_vs_lq-packages}c, specifically excludes cubes displaying characteristics of High-Quality Oxyfuel Cutting.

\begin{figure}[h]%
\centering
\includegraphics[width=0.7\textwidth]{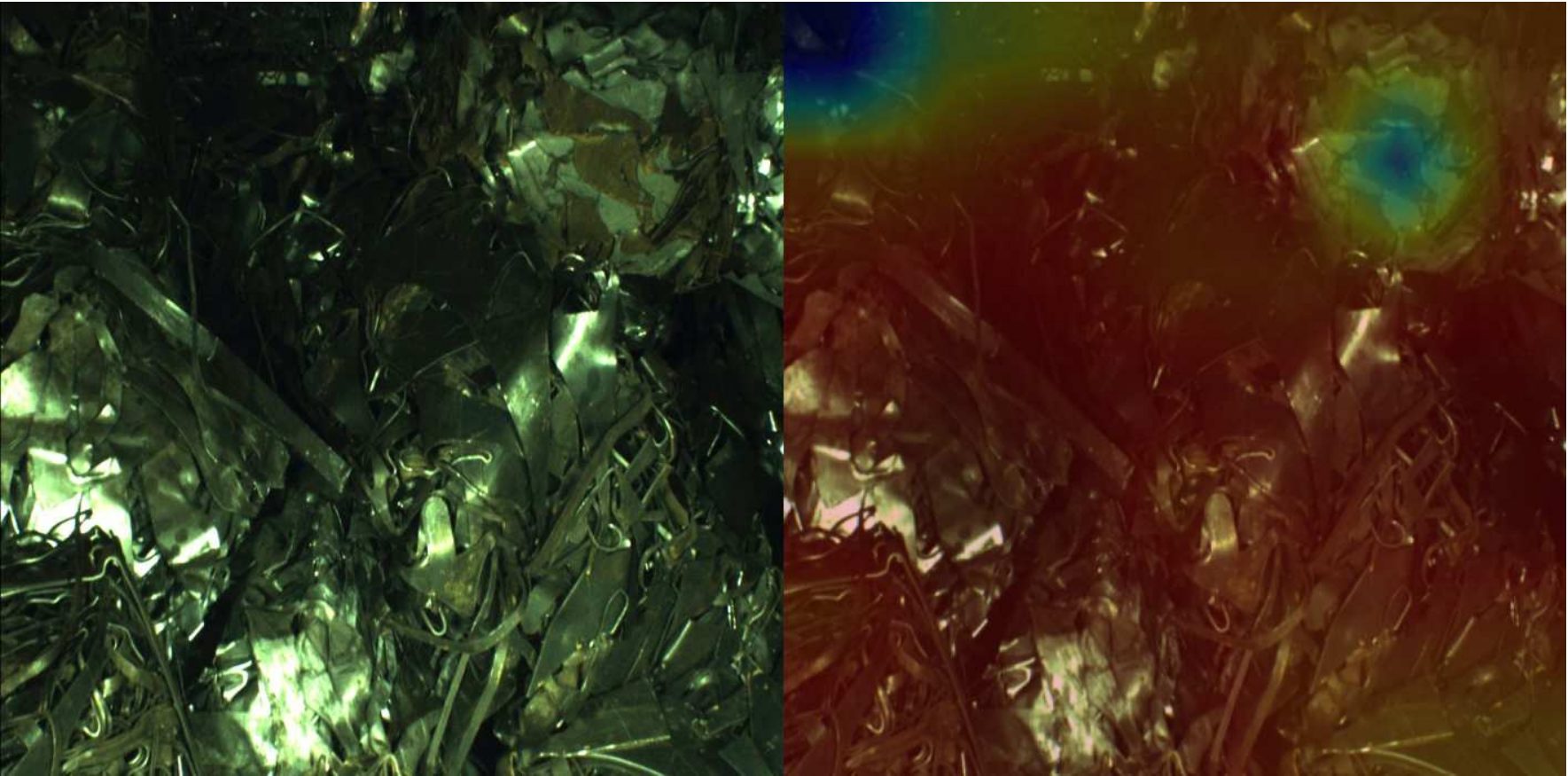}

    \begin{minipage}{0.2\textwidth}
        \centering
        (a)
    \end{minipage}
    \begin{minipage}{0.2\textwidth}
        \centering
        (b)
    \end{minipage}
    \begin{minipage}{0.2\textwidth}
        \centering
        (c)
    \end{minipage}
\caption{Understanding High-Quality Packages classification with packages with more Oxyfuel Cutting features and with Score-CAM for Swin model, \textbf{(a)} is the input, \textbf{(b)} is the heat map for High-Quality Packages. }\label{fig:hq-packages-oxy-heat map}
\end{figure}

To further examine whether the model has also learned to recognize cubes bearing High-Quality Oxyfuel Cutting features as indicative of High-Quality Packages, a sample predominantly exhibiting these features was analyzed. Figure \ref{fig:hq-packages-oxy-heat map}b reveals a heat map for High-Quality Packages showcasing a substantial area of intense heat. This pattern suggests that the model indeed leverages these features as a significant criterion for classifying High-Quality Packages. Remarkably, this sample was accurately classified, as evidenced by an $s_i$ score of 0.001, underscoring the model's adeptness at incorporating a diverse range of features to inform its classification decisions.


\begin{figure}[h]%
\centering
\includegraphics[width=0.5\textwidth]{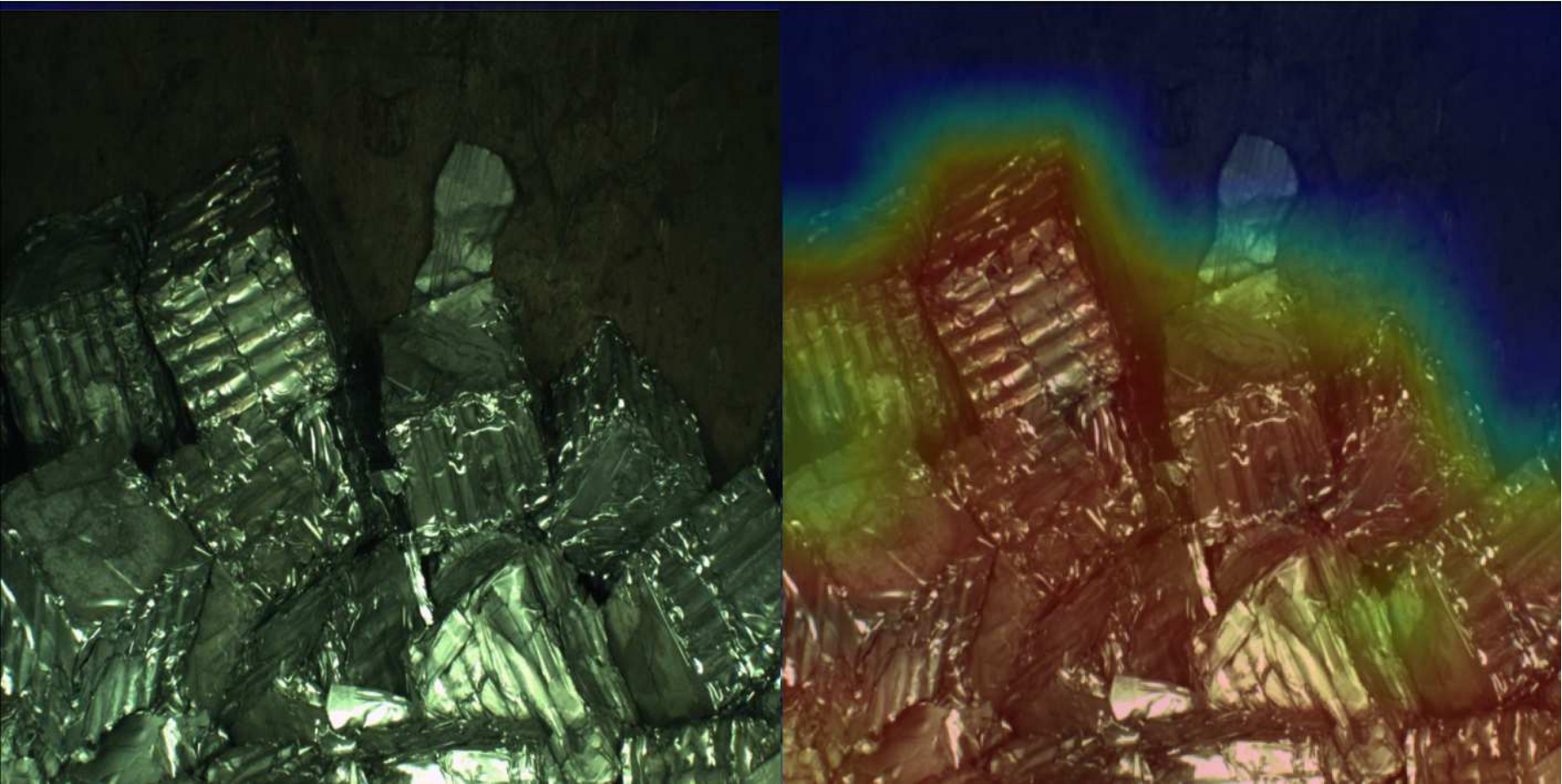}

    \begin{minipage}{0.2\textwidth}
        \centering
        (a)
    \end{minipage}
    \begin{minipage}{0.2\textwidth}
        \centering
        (b)
    \end{minipage}
    \begin{minipage}{0.2\textwidth}
        \centering
        (c)
    \end{minipage}
\caption{Understanding High-Quality Packages classification with packages with more stamping features and with Score-CAM for Swin model, \textbf{(a)} is the input, \textbf{(b)} is the heat map for High-Quality Packages. }\label{fig:hq-packages-stamp-heat map}
\end{figure}

Figure \ref{fig:hq-packages-stamp-heat map} further demonstrates that in images exclusively characterized by Stamping features, the model utilizes all the cubes present to identify High-Quality Packages. This sample was accurately classified, as indicated by an $s_i$ score of 0.006, showcasing the model's proficiency in leveraging specific visual patterns to ascertain the correct classification category.


\begin{figure}[h]%
\centering
\includegraphics[width=0.6\textwidth]{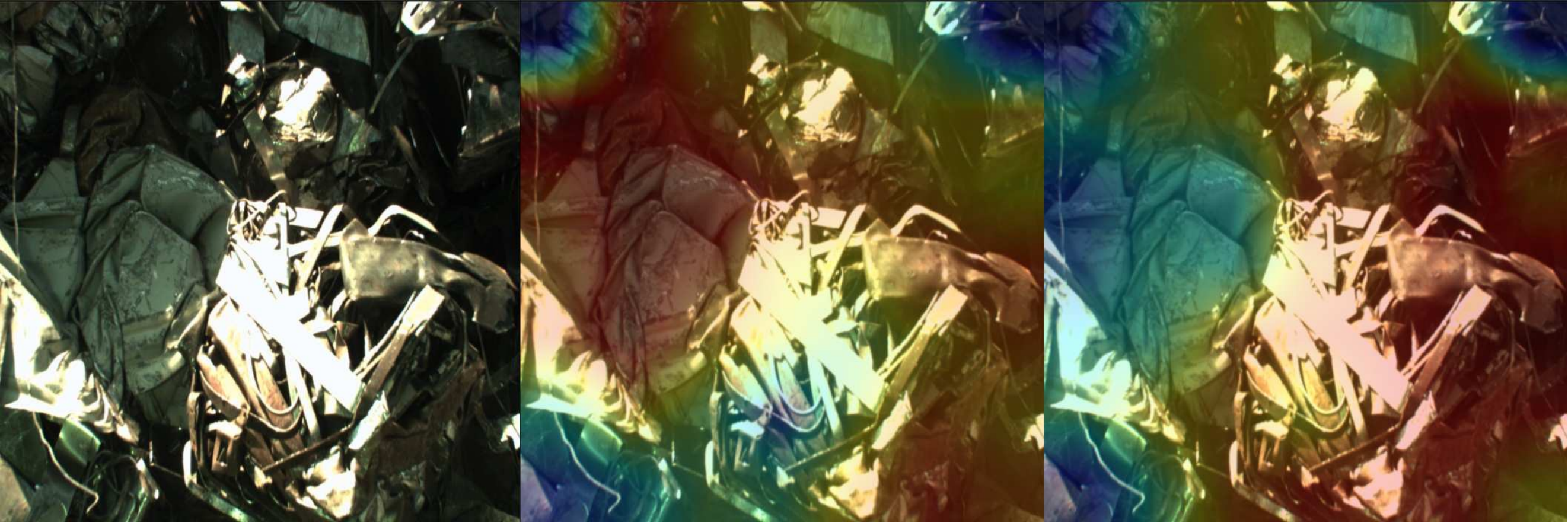}

    \begin{minipage}{0.2\textwidth}
        \centering
        (a)
    \end{minipage}
    \begin{minipage}{0.2\textwidth}
        \centering
        (b)
    \end{minipage}
    \begin{minipage}{0.2\textwidth}
        \centering
        (c)
    \end{minipage}
\caption{Understanding why the model confuses the Low-Quality Packages classification with Score-CAM for Swin model, \textbf{(a)} is the input, \textbf{(b)} is the heat map for Low-Quality Packages class, and \textbf{(c)} is the heat map for High-Quality Packages. }\label{fig:lq-packages_as_hq-packages}
\end{figure}

High-Quality Packages and Low-Quality Packages are very similar in some cases. Figure \ref{fig:lq-packages_as_hq-packages}a shows a sample of Low-Quality Packages classified as High-Quality Packages with $s_i$ as 0.28. Figure \ref{fig:lq-packages_as_hq-packages}b is the heat map for Low-Quality Packages and shows that the hot region is where there is a cube with white ink. The ink is a feature of Low-Quality Packages and is not of High-Quality Packages. Figure \ref{fig:lq-packages_as_hq-packages}c is the heat map for High-Quality Packages, and it avoids the ink appointed in Figure \ref{fig:lq-packages_as_hq-packages}b, however, there is intense light on one of the cubes, and the model sees this cube as a High-Quality Packages feature, classifying it incorrectly.    


\begin{figure}[h]%
\centering
\includegraphics[width=0.8\textwidth]{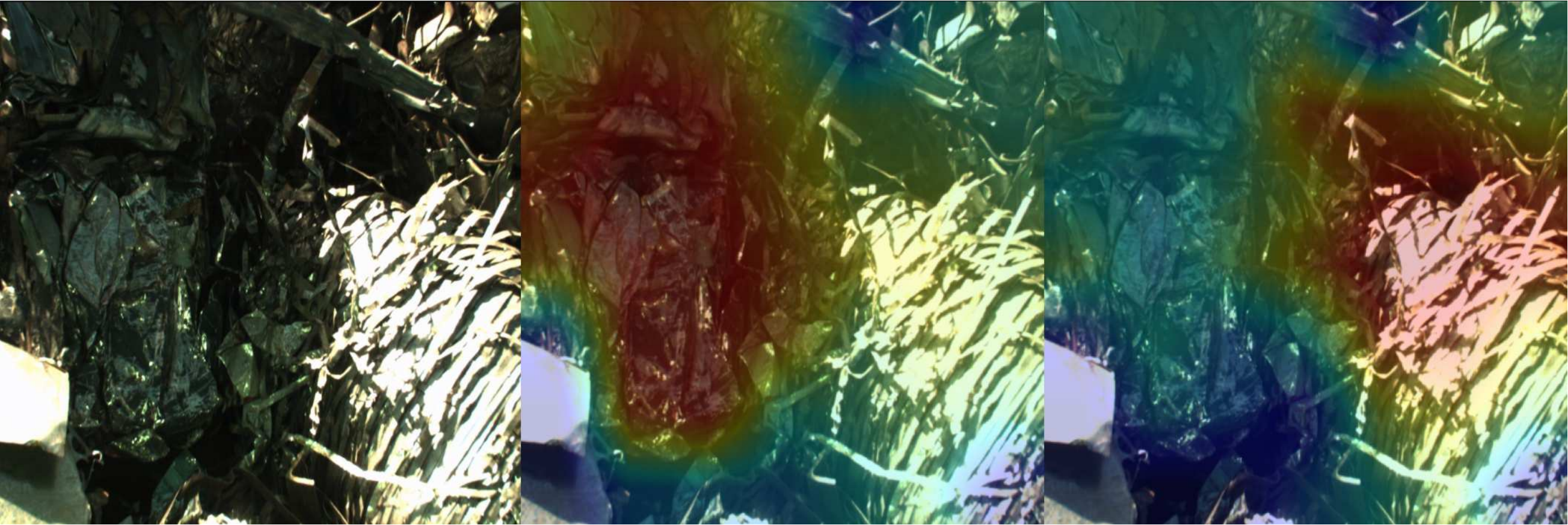}

    \begin{minipage}{0.2\textwidth}
        \centering
        (a)
    \end{minipage}
    \begin{minipage}{0.2\textwidth}
        \centering
        (b)
    \end{minipage}
    \begin{minipage}{0.2\textwidth}
        \centering
        (c)
    \end{minipage}
\caption{Understanding why the model confuses the High-Quality Packages classification with Score-CAM for Swin model, \textbf{(a)} is the input, \textbf{(b)} is the heat map for Low-Quality Packages class, and \textbf{(c)} is the heat map for High-Quality Packages. }\label{fig:hq-packages_as_lq-packages}
\end{figure}

High-Quality Packages and Low-Quality Packages can exhibit a striking resemblance under certain conditions. Figure \ref{fig:hq-packages_as_lq-packages}a presents an instance where Low-Quality Packages were erroneously classified as High-Quality Packages, evidenced by an $s_i$ score of 0.28. The heat map for Low-Quality Packages in Figure \ref{fig:hq-packages_as_lq-packages}b highlights a significant area of interest around a cube marked with white ink. This ink is a distinctive feature of Low-Quality Packages and not typically associated with High-Quality Packages. Conversely, the heat map for High-Quality Packages, depicted in Figure \ref{fig:hq-packages_as_lq-packages}c, conspicuously omits the ink-marked region identified in the preceding heat map. However, a significative illumination on one of the cubes misleads the model into recognizing it as a characteristic of High-Quality Packages, leading to incorrect classification. 


\begin{figure}[h]%
\centering
\includegraphics[width=0.8\textwidth]{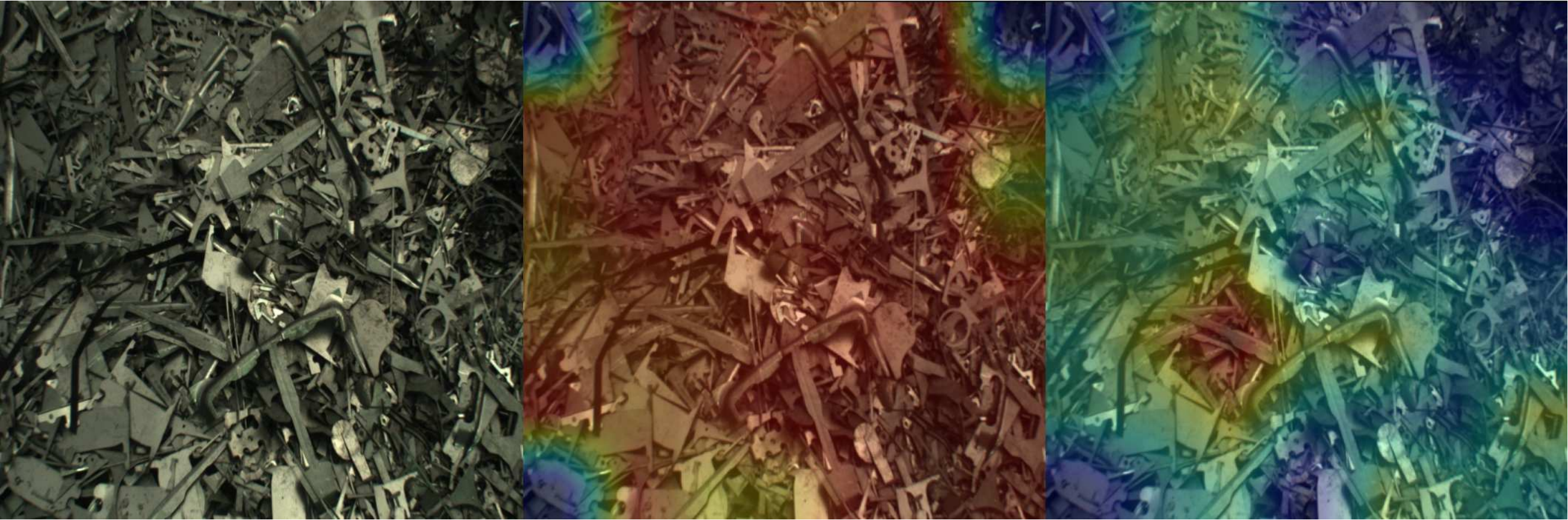}

    \begin{minipage}{0.2\textwidth}
        \centering
        (a)
    \end{minipage}
    \begin{minipage}{0.2\textwidth}
        \centering
        (b)
    \end{minipage}
    \begin{minipage}{0.2\textwidth}
        \centering
        (c)
    \end{minipage}
\caption{Understanding Stamping Scrap classification with Score-CAM for Swin model, \textbf{(a)} is the input, \textbf{(b)} is the heat map for Stamping Scrap, and \textbf{(c)} is the heat map for High-Quality Oxyfuel Cutting. }\label{fig:stamping_it_vs_hq-oxyfuel}
\end{figure}

Stamping Scrap and High-Quality Oxyfuel Cutting represent high-quality classes, distinguished by their lack of contaminants, with differentiation primarily based on shape and color. Figure \ref{fig:stamping_it_vs_hq-oxyfuel}a displays a correctly classified stamping scrap sample, as evidenced by a low $s_i$ score of 0.0001. The heat map for Stamping, shown in Figure \ref{fig:stamping_it_vs_hq-oxyfuel}b, reveals a significantly active region, indicating the model's strong association of these visual cues with the Stamping classification. Conversely, the heat map for High-Quality Oxyfuel Cutting, illustrated in Figure \ref{fig:stamping_it_vs_hq-oxyfuel}c, remains notably inactive, suggesting the model's discernment in distinguishing between the two classes based on the available visual indicators. 


\begin{figure}[h]%
\centering
\includegraphics[width=0.8\textwidth]{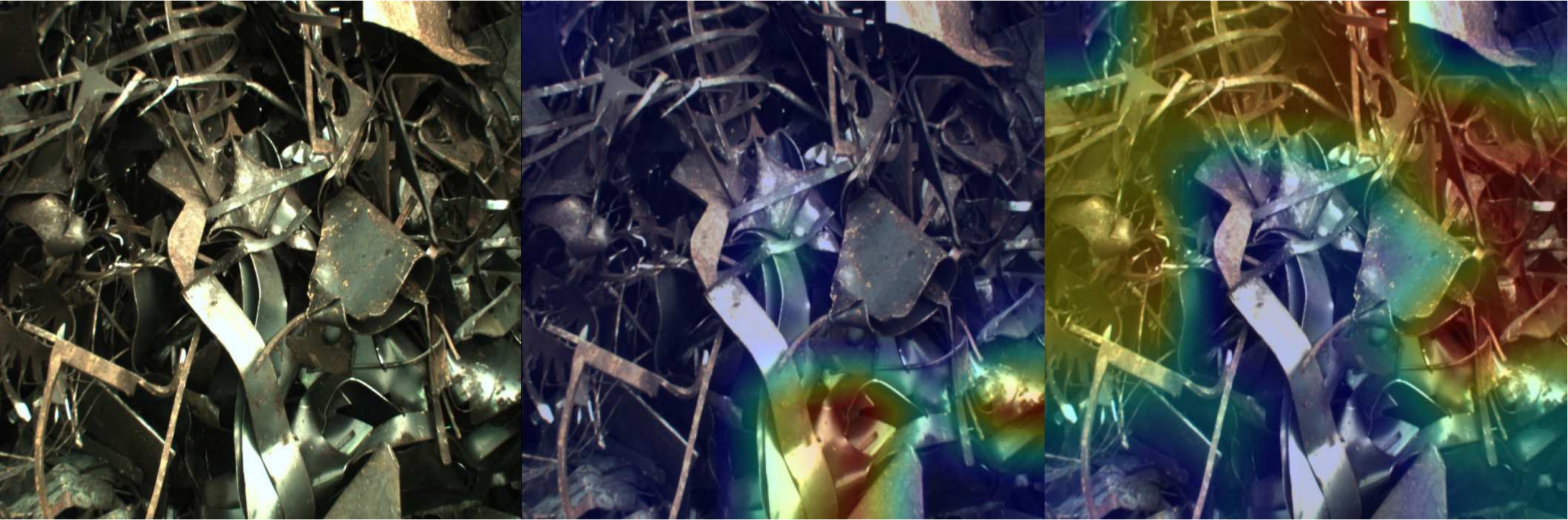}

    \begin{minipage}{0.2\textwidth}
        \centering
        (a)
    \end{minipage}
    \begin{minipage}{0.2\textwidth}
        \centering
        (b)
    \end{minipage}
    \begin{minipage}{0.2\textwidth}
        \centering
        (c)
    \end{minipage}
\caption{Understanding High-Quality Oxyfuel Cutting classification with Score-CAM for Swin model, \textbf{(a)} is the input, \textbf{(b)} is the heat map for Stamping Scrap, and \textbf{(c)} is the heat map for High-Quality Oxyfuel Cutting. }\label{fig:hq-oxyfuel_it_vs_stamping}
\end{figure}

An illustrative example of High-Quality Oxyfuel Cutting is presented in Figure \ref{fig:hq-oxyfuel_it_vs_stamping}a, where the sample was accurately classified with an $s_i$ score of 0.003. The heat map for Stamping, depicted in Figure \ref{fig:hq-oxyfuel_it_vs_stamping}b, highlights a limited area showcasing a bright silver hue, a feature more commonly associated with Stamping Scrap, as illustrated in Figure \ref{fig:stamping_it_vs_hq-oxyfuel}b. Conversely, the heat map for High-Quality Oxyfuel Cutting, seen in Figure \ref{fig:hq-oxyfuel_it_vs_stamping}b, is more pronounced in regions exhibiting a rust-like coloration. 


\begin{figure}[h]%
\centering
\includegraphics[width=0.8\textwidth]{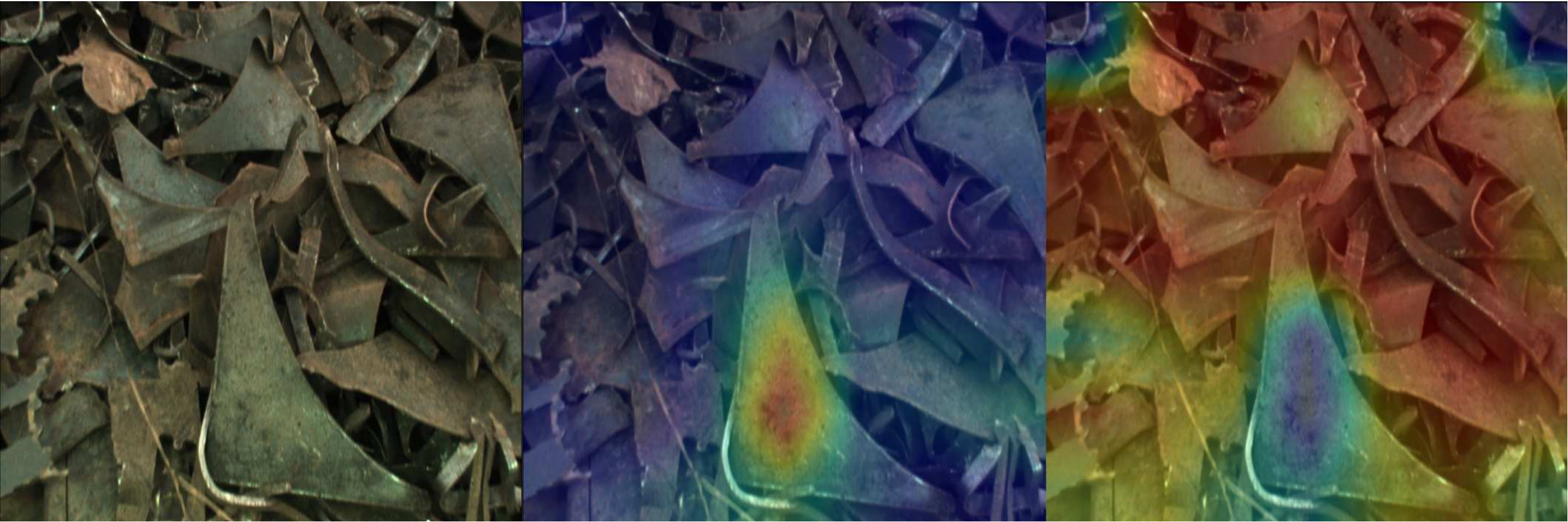}

    \begin{minipage}{0.2\textwidth}
        \centering
        (a)
    \end{minipage}
    \begin{minipage}{0.2\textwidth}
        \centering
        (b)
    \end{minipage}
    \begin{minipage}{0.2\textwidth}
        \centering
        (c)
    \end{minipage}
\caption{Understanding High-Quality Oxyfuel Cutting classification with Score-CAM for Swin model, \textbf{(a)} is the input, \textbf{(b)} is the heat map for Low-Quality Oxyfuel Cutting, and \textbf{(c)} is the heat map for High-Quality Oxyfuel Cutting. }\label{fig:hq-oxyfuel_it_vs_lq-oxyfuel}
\end{figure}

High-Quality and Low-Quality Oxyfuel Cutting Scraps, despite sharing the same cutting process, are distinguished by their raw material composition. The presence of ink is a defining characteristic of the Low-Quality Oxyfuel Cutting class. Figure \ref{fig:hq-oxyfuel_it_vs_lq-oxyfuel}a displays a correctly classified example of High-Quality Oxyfuel Cutting, with an $s_i$ score of 0.003. Figure \ref{fig:hq-oxyfuel_it_vs_lq-oxyfuel}b presents a muted heat map for Low-Quality Oxyfuel Cutting, contrasting with the more active heat map for High-Quality Oxyfuel Cutting in Figure \ref{fig:hq-oxyfuel_it_vs_lq-oxyfuel}c.


\begin{figure}[h]%
\centering
\includegraphics[width=0.8\textwidth]{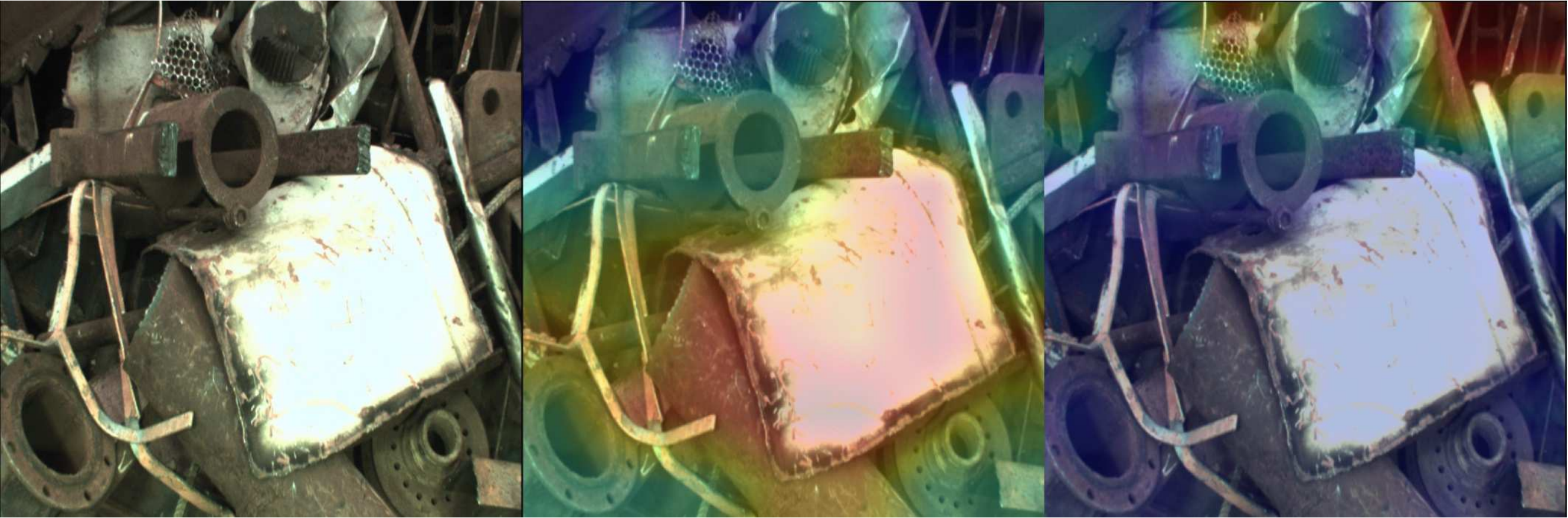}

    \begin{minipage}{0.2\textwidth}
        \centering
        (a)
    \end{minipage}
    \begin{minipage}{0.2\textwidth}
        \centering
        (b)
    \end{minipage}
    \begin{minipage}{0.2\textwidth}
        \centering
        (c)
    \end{minipage}
\caption{Understanding Low-Quality Oxyfuel Cutting classification with Score-CAM for Swin model, \textbf{(a)} is the input, \textbf{(b)} is the heat map for Low-Quality Oxyfuel Cutting, and \textbf{(c)} is the heat map for High-Quality Oxyfuel Cutting. }\label{fig:lq-oxyfuel_it_vs_hq-oxyfuel}
\end{figure}

Conversely, Figure \ref{fig:lq-oxyfuel_it_vs_hq-oxyfuel}a features a sample of Low-Quality Oxyfuel Cutting correctly identified with an $s_i$ score of 0.0009. Here, the heat map for Low-Quality Oxyfuel Cutting in Figure \ref{fig:lq-oxyfuel_it_vs_hq-oxyfuel}b highlights a significant area of interest where ink is present on the scrap. Meanwhile, the heat map for High-Quality Oxyfuel Cutting in Figure \ref{fig:lq-oxyfuel_it_vs_hq-oxyfuel}c remains subdued, clearly delineating regions containing ink.


\begin{figure}[h]%
\centering
\includegraphics[width=0.6\textwidth]{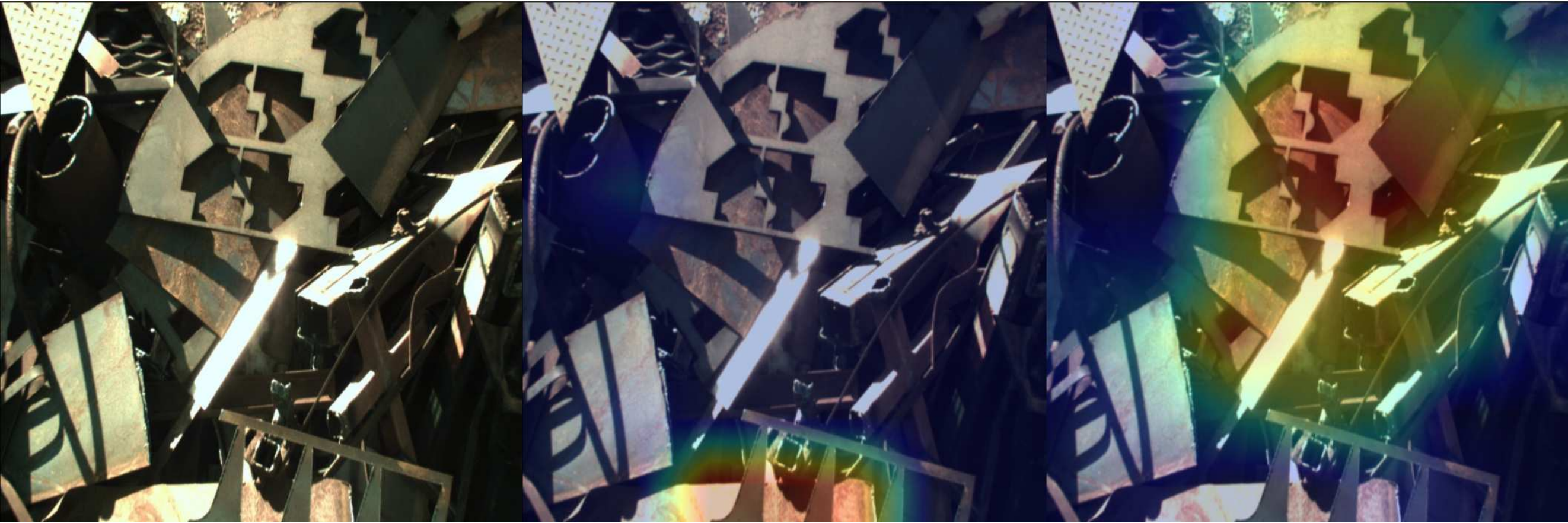}

    \begin{minipage}{0.2\textwidth}
        \centering
        (a)
    \end{minipage}
    \begin{minipage}{0.2\textwidth}
        \centering
        (b)
    \end{minipage}
    \begin{minipage}{0.2\textwidth}
        \centering
        (c)
    \end{minipage}
\caption{Understanding why the model confuses the Low-Quality Oxyfuel Cutting classification with Score-CAM for Swin model, \textbf{(a)} is the input, \textbf{(b)} is the heat map for Low-Quality Oxyfuel Cutting, and \textbf{(c)} is the heat map for High-Quality Oxyfuel Cutting. }\label{fig:lq-oxyfuel_as_hq-oxyfuel}
\end{figure}

Figure \ref{fig:lq-oxyfuel_as_hq-oxyfuel}a provides an instance where the model incorrectly identifies Low-Quality Oxyfuel Cutting as High-Quality Oxyfuel Cutting, assigning an $s_i$ score of 0.42. Despite exhibiting characteristics of lower quality, the absence of ink or other noticeable contaminants in this sample likely contributed to the model's misclassification. The heat map for Low-Quality Oxyfuel Cutting, visible in Figure \ref{fig:lq-oxyfuel_as_hq-oxyfuel}b, shows minimal activity, whereas the heat map for High-Quality Oxyfuel Cutting in Figure \ref{fig:lq-oxyfuel_as_hq-oxyfuel}c reveals a more pronounced area of focus. This suggests that the model may erroneously classify samples as higher quality in the absence of explicit low-quality indicators such as ink.

The subsequent examples illustrate the classes Shredder, Steel Sheets, and Swarf, where the models demonstrated exemplary recall rates. Notably, both the Swin and ViT models correctly classified all samples from these categories.

\begin{figure}[h]%
\centering
\includegraphics[width=0.5\textwidth]{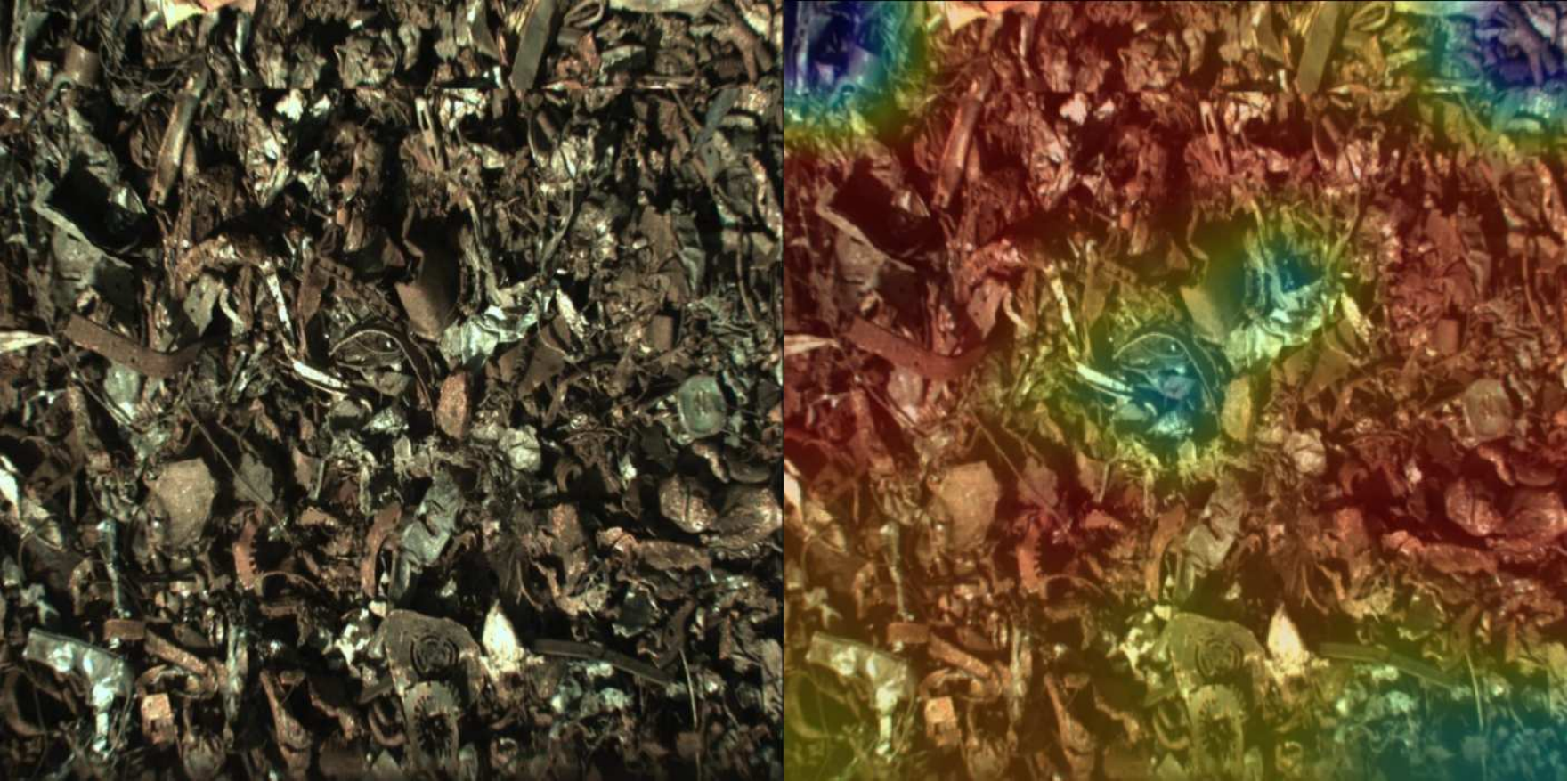}

    \begin{minipage}{0.2\textwidth}
        \centering
        (a)
    \end{minipage}
    \begin{minipage}{0.2\textwidth}
        \centering
        (b)
    \end{minipage}
\caption{Understanding Shredder classification with Score-CAM for Swin model, \textbf{(a)} is the input, \textbf{(b)} is the heat map for Shredder.}\label{fig:shredder_1}
\end{figure}

\begin{figure}[h]%
\centering
\includegraphics[width=0.8\textwidth]{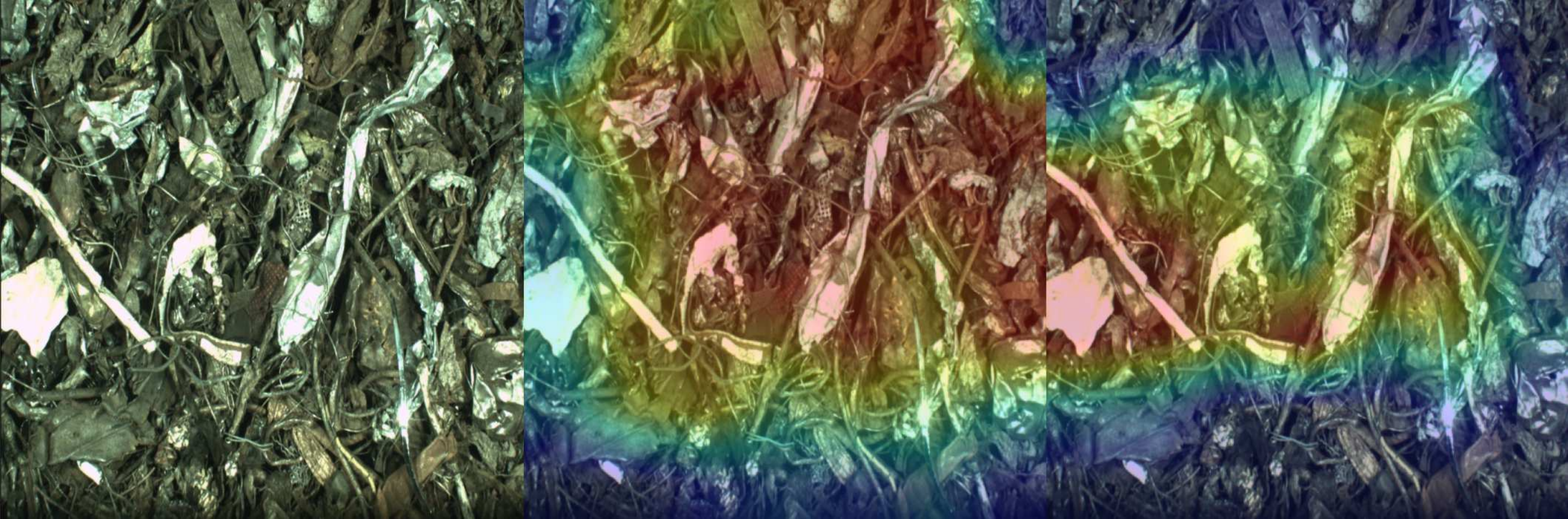}

    \begin{minipage}{0.2\textwidth}
        \centering
        (a)
    \end{minipage}
    \begin{minipage}{0.2\textwidth}
        \centering
        (b)
    \end{minipage}
    \begin{minipage}{0.2\textwidth}
        \centering
        (c)
    \end{minipage}
\caption{Understanding Shredder classification with Score-CAM for Swin model, \textbf{(a)} is the input, \textbf{(b)} is the heat map for Shredder, and \textbf{(c)} is the heat map for Sheared Scrap. }\label{fig:shredder_vs_shared}
\end{figure}

Figure \ref{fig:shredder_1} presents a Shredder sample, which is characterized by a significantly active area in the heat map. While Shredder scrap may exhibit crumpled features akin to Sheared Scrap, its pieces are markedly smaller. Figure \ref{fig:shredder_vs_shared}a offers another Shredder sample, with Figure \ref{fig:shredder_vs_shared}b highlighting Shredder characteristics and Figure \ref{fig:shredder_vs_shared}c focusing on Sheared Scrap. The Shredder heat map displays a more extensive area of interest compared to Sheared. However, the Sheared heat map suggests that crumpled aspects are pertinent to Shredder classification as well.

\begin{figure}[h]%
\centering
\includegraphics[width=0.7\textwidth]{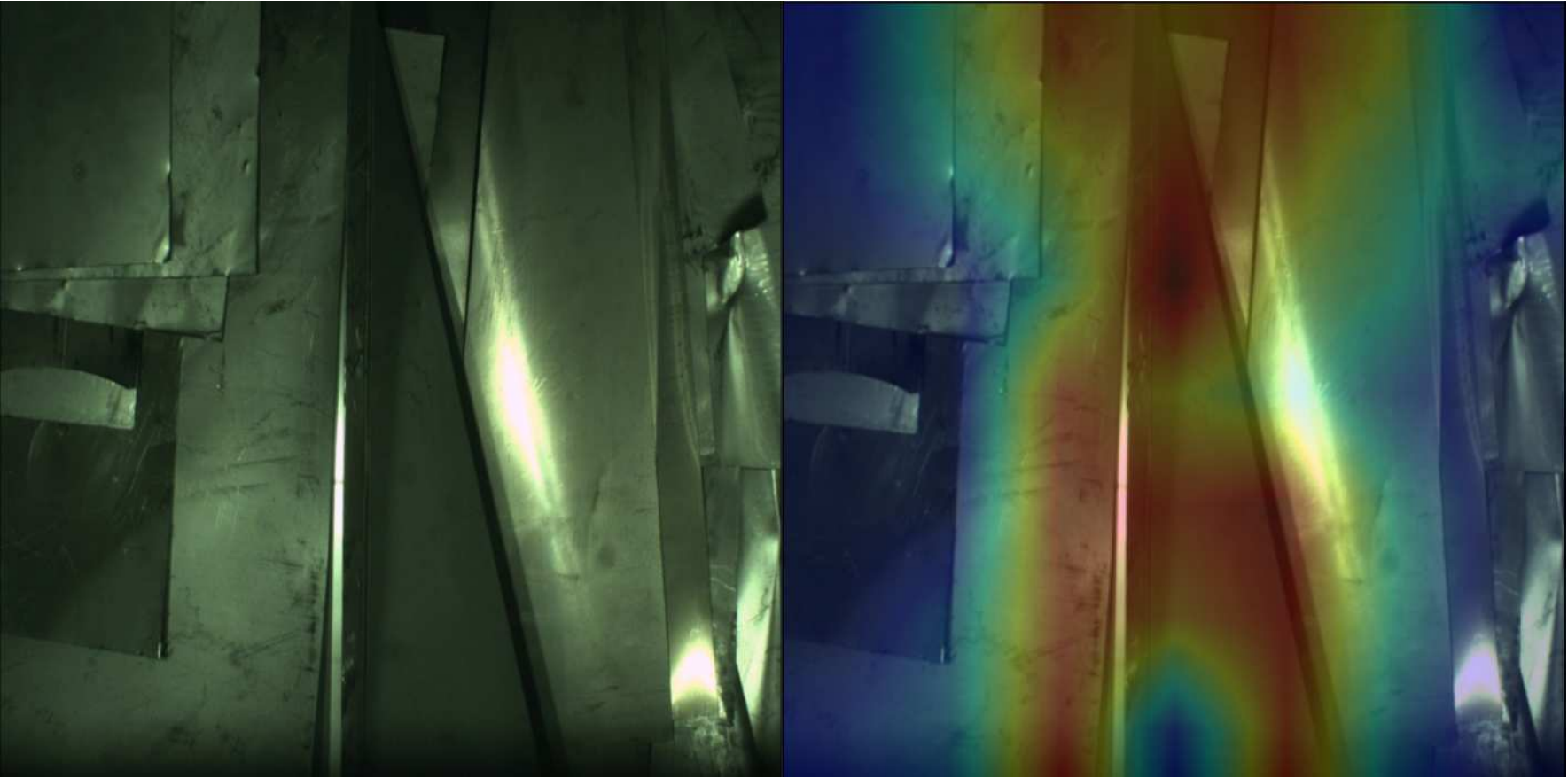}

    \begin{minipage}{0.2\textwidth}
        \centering
        (a)
    \end{minipage}
    \begin{minipage}{0.2\textwidth}
        \centering
        (b)
    \end{minipage}
\caption{Understanding Steel Sheets classification with Score-CAM for Swin model, \textbf{(a)} is the input, \textbf{(b)} is the heat map for Steel Sheets.}\label{fig:steel_sheets_1}
\end{figure}

\begin{figure}
\centering
\includegraphics[width=0.6\textwidth]{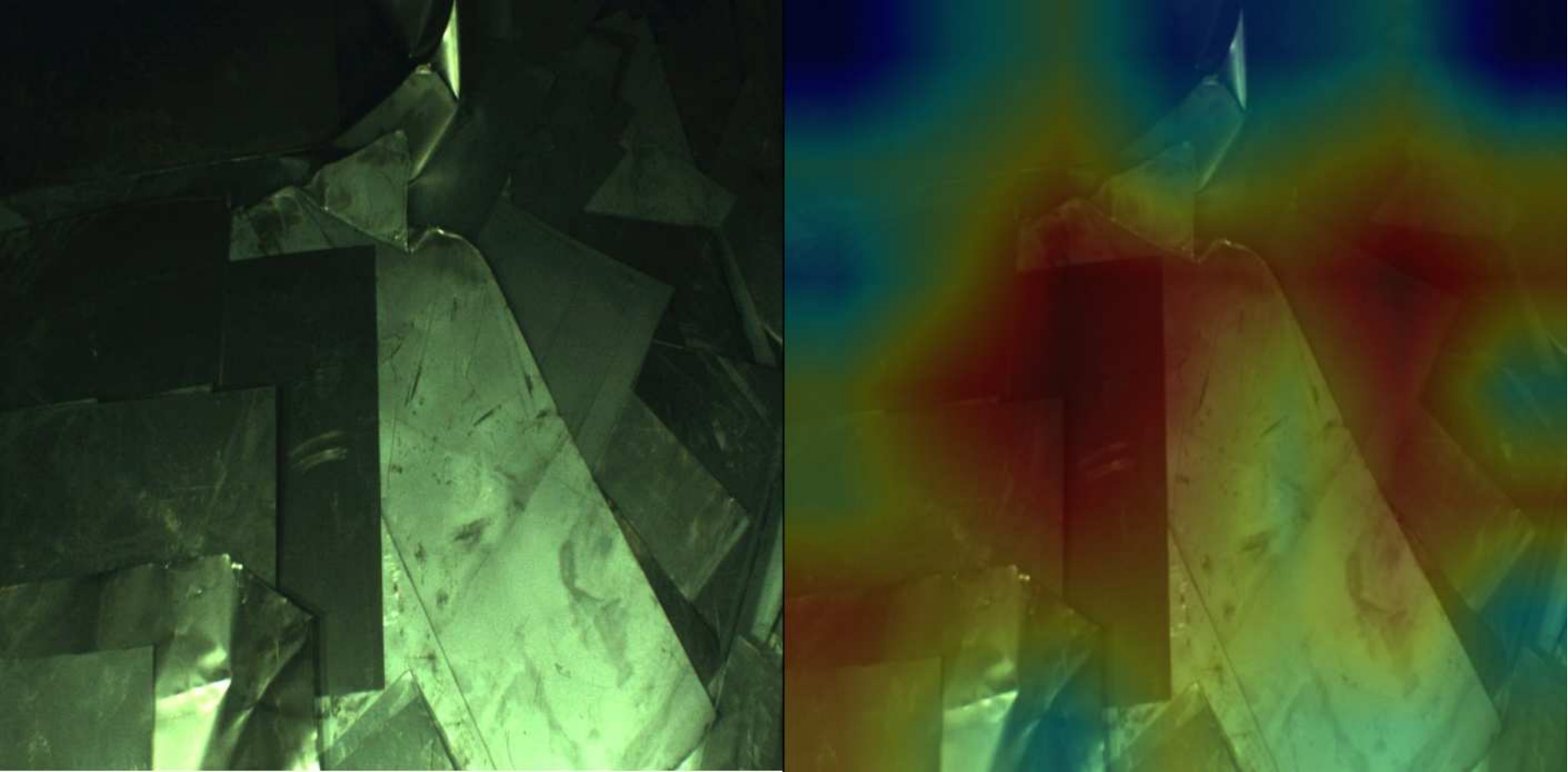}

    \begin{minipage}{0.2\textwidth}
        \centering
        (a)
    \end{minipage}
    \begin{minipage}{0.2\textwidth}
        \centering
        (b)
    \end{minipage}
\caption{Understanding Steel Sheets classification with Score-CAM for Swin model, \textbf{(a)} is the input, \textbf{(b)} is the heat map for Steel Sheets.}\label{fig:steel_sheets_2}
\end{figure}

Regarding Steel Sheets, Figures \ref{fig:steel_sheets_1} and \ref{fig:steel_sheets_2} reveal that the model prioritizes the edges of the sheets as key features for classification, underscoring the geometric precision in identifying this class.

\begin{figure}
\centering
\includegraphics[width=0.6\textwidth]{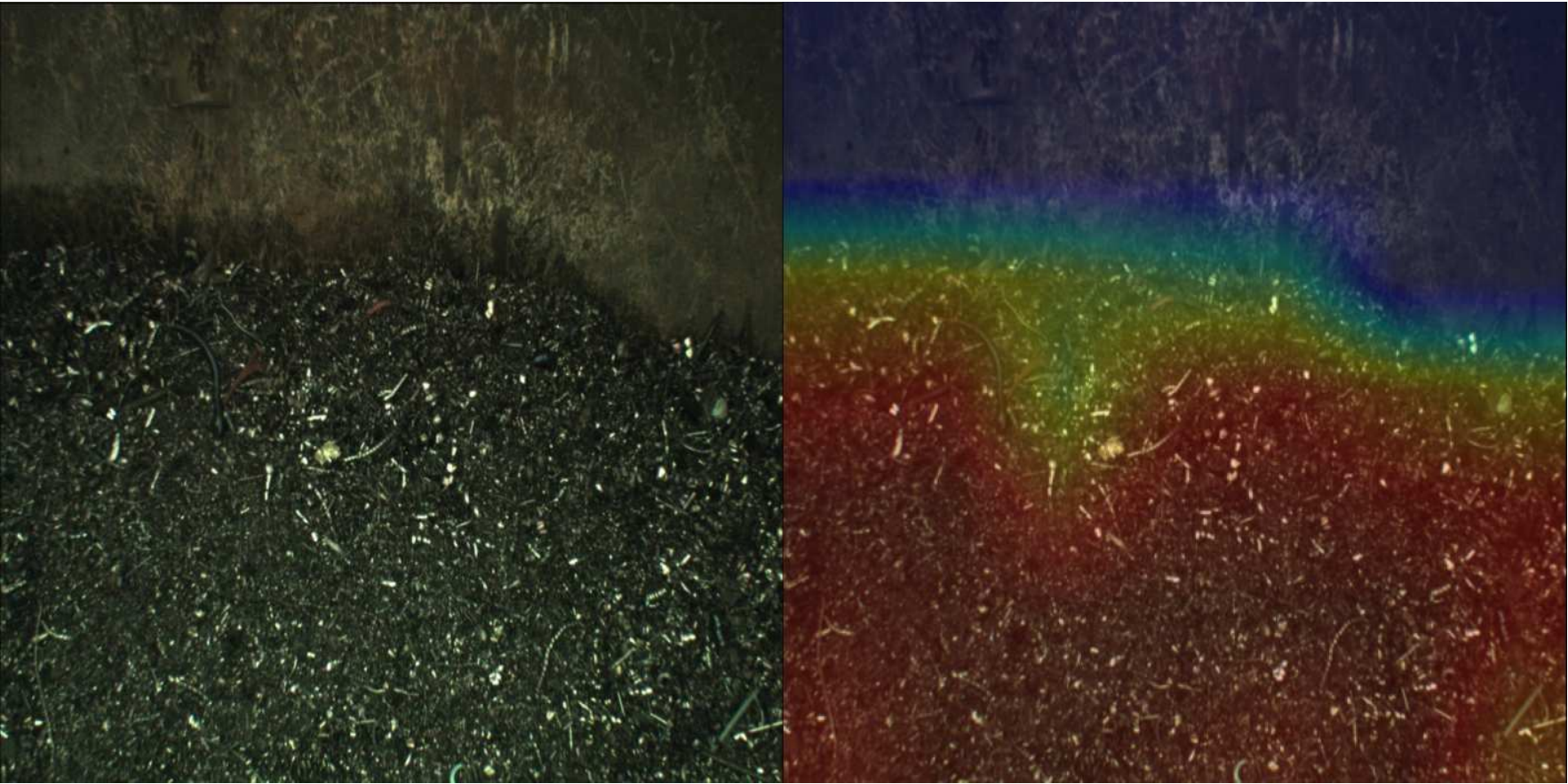}

    \begin{minipage}{0.2\textwidth}
        \centering
        (a)
    \end{minipage}
    \begin{minipage}{0.2\textwidth}
        \centering
        (b)
    \end{minipage}
\caption{Understanding Swarf Scrap classification with Score-CAM for Swin model, \textbf{(a)} is the input, \textbf{(b)} is the heat map for Swarf Scrap.}\label{fig:swarf_1}
\end{figure}

\begin{figure}
\centering
\includegraphics[width=0.6\textwidth]{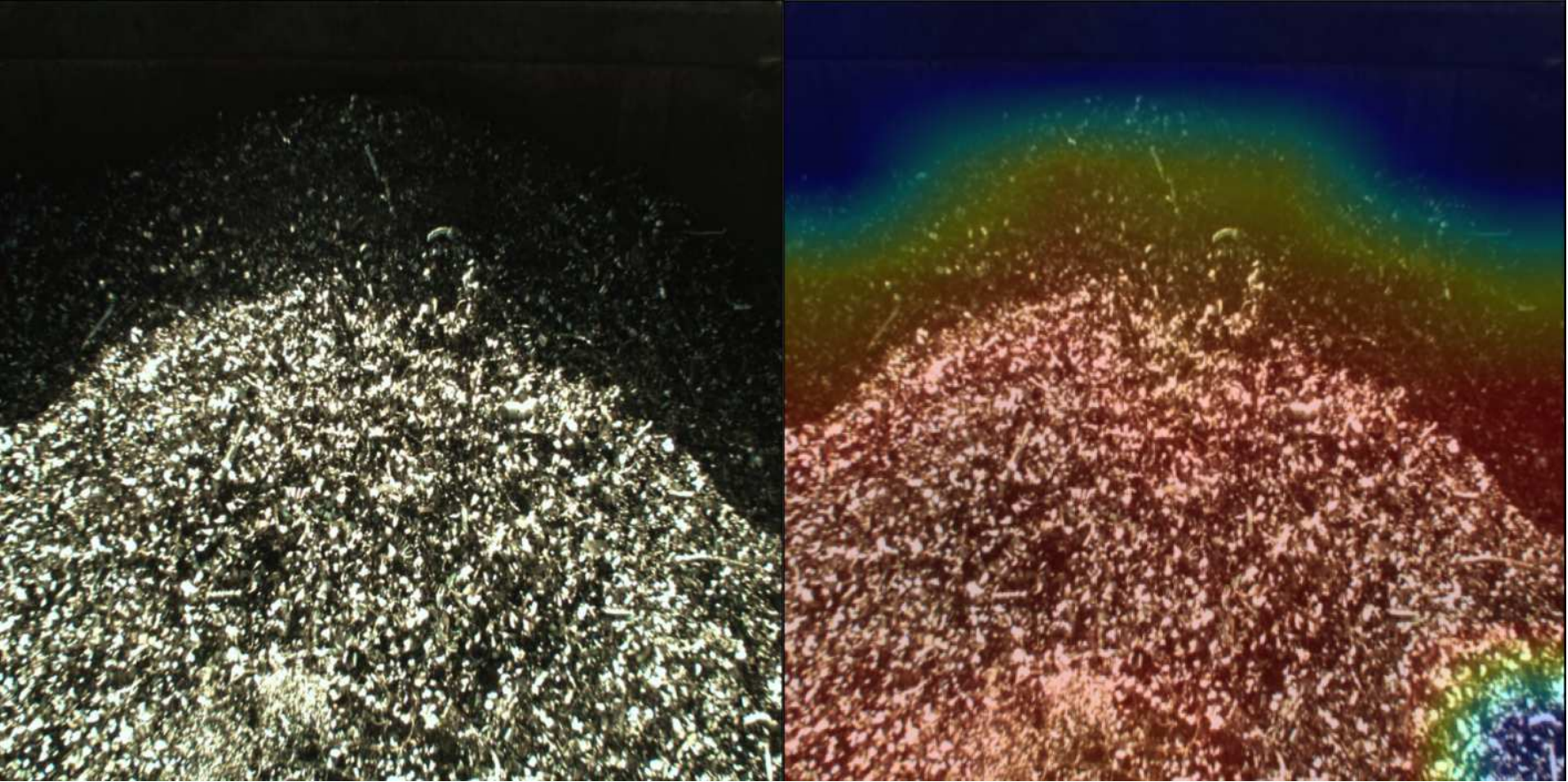}

    \begin{minipage}{0.2\textwidth}
        \centering
        (a)
    \end{minipage}
    \begin{minipage}{0.2\textwidth}
        \centering
        (b)
    \end{minipage}
\caption{Understanding Swarf Scrap classification with Score-CAM for Swin model, \textbf{(a)} is the input, \textbf{(b)} is the heat map for Swarf Scrap.}\label{fig:swarf_2}
\end{figure}

Swarf Scrap is distinguished by its unique shape and color, which sets it apart from other categories. Figures \ref{fig:swarf_1} and \ref{fig:swarf_2} depict Swarf Scrap samples, with the heat maps illustrating extensive areas of interest and effectively segmenting scrap from non-scrap elements. Intriguingly, despite the presence of shadows in Figure \ref{fig:swarf_2}, the heat map remains intensely focused on the scrap-containing regions, indicating the model's adeptness at recognizing Swarf Scrap's distinctive attributes even in less-than-ideal lighting conditions.

\section{Conclusions}\label{sec4}

In this study, we examined the classification of ferrous scrap materials with an emphasis on explicability and quantification of inference uncertainties, guided by two central research questions. The first question asked how the integration of Split Conformal Prediction with advanced deep learning models can enhance the certainty and reliability of classification outcomes in industrial applications, particularly for ferrous scrap classification. The second question explored how various explainability methods could elucidate the decision-making processes of these models, especially in identifying features that deviate from typical class characteristics. 

To address these inquiries, we constructed a comprehensive database of 8,147 images spanning nine distinct classes of scrap materials. Our experimental framework employed three sophisticated deep learning network architectures: ResNet-50, Vision Transformer (ViT), and Swin Transformer. Each model's ability to manage uncertainties and provide transparent, interpretable decisions was rigorously tested, enabling us to assess the efficacy of conformal prediction techniques in selecting the optimal model and to determine the impact of explainability metrics on model selection and classification accuracy.

The findings revealed that the models achieved high average accuracy rates: 95.00\% for Resnet-50, 95.15\% for ViT, and 95.51\% for Swin. Furthermore, the application of the Split Conformal Prediction method allowed for the quantification of each model's uncertainties, enhancing the understanding of predictions and increasing the reliability of the results. Specifically, the Swin model demonstrated more reliable outcomes than the others, as evidenced by the smaller average size of prediction sets. In addition, Swin showed the shortest average training time, approximately 53 minutes. 

During the analysis of the results, it was noted that the models often confused the "Sheared Scrap" class with "Low-Quality Oxyfuel Cutting", both made of the same material but processed differently. The explicability provided by the Score-CAM method, in conjunction with the Swin model, allowed for the identification of key scrap characteristics, including the presence of contaminants such as paint, and distinguished between formats. Furthermore, some samples labeled as "Sheared Scrap" were found by the Score-CAM heat map to exhibit strong characteristics of "Low-Quality Oxyfuel Cutting", such as burn marks from Oxyfuel cutting, highlighting the challenges in labeling some images. Overall, this study presented a reliable and explainable method for the classification of ferrous scraps, addressing gaps in the literature and providing valuable insights into the application of explainability methods and uncertainty quantification in artificial intelligence models for this purpose. In addition, the findings have significant practical implications for the steel industry, where efficient scrap classification is crucial for optimizing production processes, ensuring final product quality, and contributing to environmental sustainability.

Despite the promising results of this study, it is important to acknowledge certain limitations that may have influenced the outcomes. These include challenges in accurately labeling the dataset due to difficulties in distinguishing some classes, even for human experts, and the potential for noise introduction in model training by images containing multiple classes. The limited number of samples in the dataset also poses a constraint, underscoring the need for dataset expansion and diversification to enhance model accuracy and generalization.

\section*{Acknowledgments}

The authors would also like to thank the \textit{Coordenação de Aperfeiçoamento de Pessoal de Nível Superior} - Brazil (CAPES) - Finance Code 001, \textit{Fundacão de Amparo à Pesquisa do Estado de Minas Gerais} (FAPEMIG, grants APQ-01518-21), \textit{Conselho Nacional de Desenvolvimento Científico e Tecnológico} (CNPq, grants 308400/2022-4), the Instituto Tecnológico Vale (ITV)  and Universidade Federal de Ouro Preto (PROPPI/UFOP) for supporting the development of this study.

\bibliography{sn-bibliography}

\end{document}